\documentclass[runningheads]{llncs}

\usepackage{eccv}

\usepackage{eccvabbrv}

\usepackage{graphicx}
\usepackage{booktabs}
\usepackage{arydshln}
\usepackage{wrapfig}
\usepackage{afterpage}
\usepackage{xcolor}
\definecolor{warpgreen}{RGB}{106,168,79}

\usepackage[accsupp]{axessibility}  

\usepackage{hyperref}

\usepackage{orcidlink}

\begin{document}

\title{WarpI2I: Image Warping for Image-to-Image Translation} 

\titlerunning{WarpI2I}

\author{Shen Zheng \and
Anurag Ghosh \and
Gaurav Parmar \and
Srinivasa Narasimhan
}

\authorrunning{Zheng et al.}

\institute{Carnegie Mellon University}

\maketitle

\begin{abstract}

Image-to-image (I2I) translation has achieved strong results in tasks like human relighting and driving scene translation using latent diffusion models (LDMs). However, compact LDMs often struggle to preserve fine-grained structures because the encoder compresses high-resolution inputs into a spatially downsampled latent space. To address this issue, we propose a simple saliency-guided warp–unwarp framework that reallocates spatial representation toward salient regions before encoding, enabling better preservation of structural details without increasing latent resolution. The warped image is processed by the original diffusion model and then mapped back via an inverse warp. In addition, we propose a simple and efficient outpainting-based synthetic data generation pipeline to produce high-quality paired data for image relighting. Our method is model-agnostic, requires no architectural modification, and introduces negligible computational overhead. Experiments on human relighting, driving scene relighting, and translation demonstrate improved structural preservation, lighting faithfulness, and image quality, with our framework extending naturally to video via frame-by-frame application with good temporal stability. Project Webpage: \href{https://shenzheng2000.github.io/WarpI2I.github.io/}{\textcolor{eccvblue}{https://shenzheng2000.github.io/WarpI2I.github.io/}}

  \keywords{Image Warping \and Image-to-Image Translation \and Relighting}
\end{abstract}

\begin{figure}[t]
    \centering
    \includegraphics[width=\linewidth]{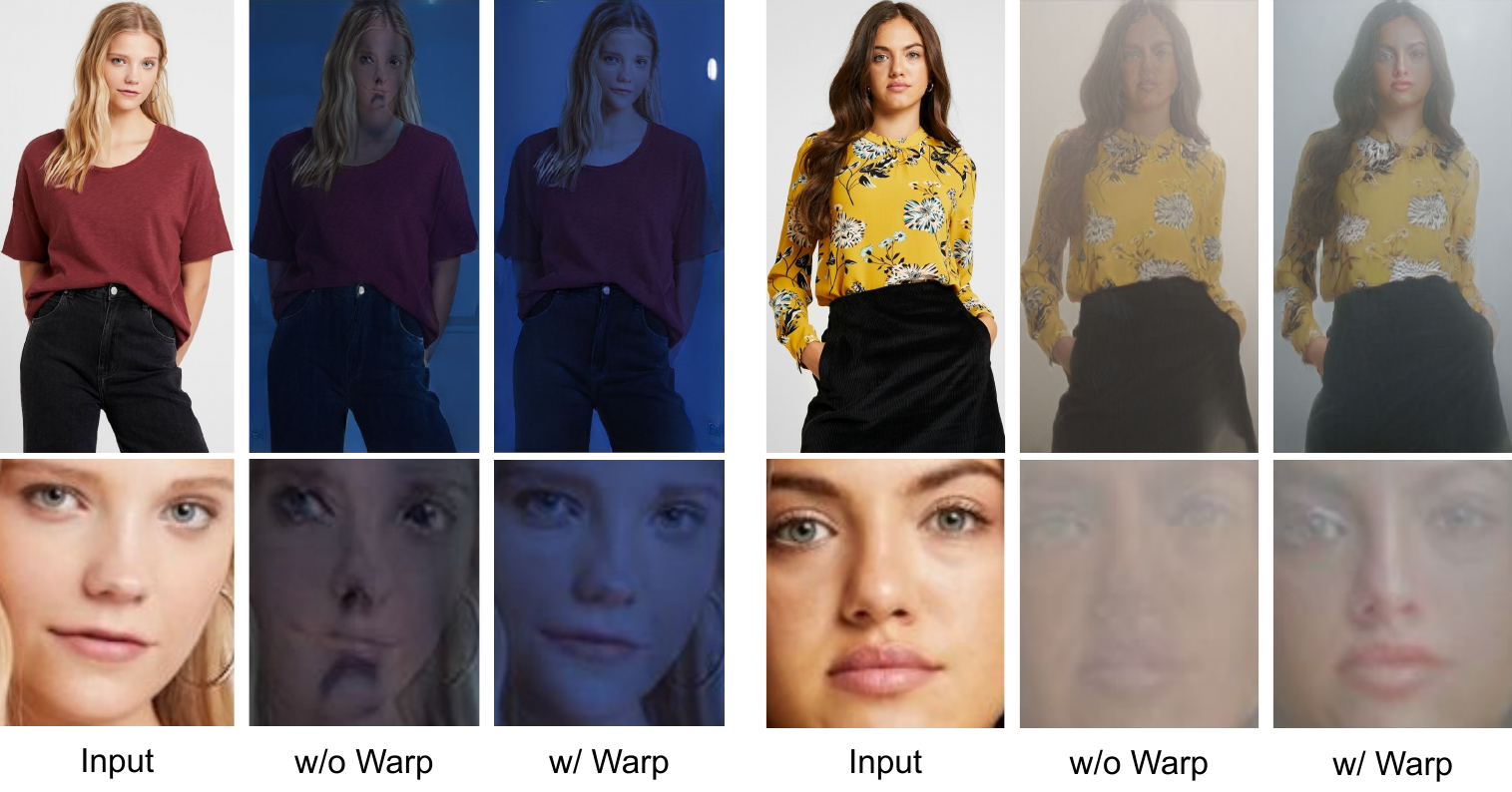}
    \vspace{-0.6cm}
    \caption{\textbf{Our warping significantly improves fine details of face regions, compared to the baseline img2img-turbo~\cite{parmar2024one}.} Real human image is from VITON-HD~\cite{choi2021viton} dataset using moonlight (\textit{left}) and foggy (\textit{right}) as prompts for human relighting.}
   \label{fig:teaser}
\end{figure}

\section{Introduction}
\label{sec:intro}



    

Image-to-image (I2I) translation has achieved remarkable progress in tasks such as human relighting and driving scene weather transformation. Latent diffusion models (LDMs)~\cite{rombach2022high,podell2023sdxl,esser2024scaling} have emerged as the dominant paradigm due to their strong photorealism, stable optimization, and scalability.

While high-capacity commercial systems (e.g., Midjourney~\cite{midjourney2024}, DALL·E 3~\cite{betker2023improving}) demonstrate impressive performance, they are closed-source and computationally intensive. Thus, lightweight open-source LDMs like Stable Diffusion~\cite{rombach2022high} and its scaled variants~\cite{podell2023sdxl, esser2024scaling} are widely adopted for practical deployment.

However, compact LDMs often struggle to preserve fine-grained foreground structures, particularly facial details and small objects. This limitation stems from the encoder’s spatial compression, which maps high-resolution inputs into a downsampled latent space and discards high-frequency structural information.

A straightforward remedy is to decrease the encoder’s compression ratio or increase the input image resolution to obtain a larger latent spatial size. However, this strategy significantly increases computational memory and inference time, as the computational cost scales quadratically with latent spatial size, making it impractical for real-world deployment. Alternative strategies such as multi-scale diffusion pipelines~\cite{ho2022cascaded, saharia2022photorealistic,bar2023multidiffusion} or test-time refinement~\cite{khan2025test,dalal2025constructive} have been explored, but these approaches typically introduce additional computational overhead during training or inference, and often require architecture modification.

In this work, instead of enlarging the latent space, we preserve the original encoder compression ratio and input resolution, and counter the spatial shrinking caused by the encoder using a content-aware spatial image warping before encoding. This warping locally enlarges salient (important) regions, giving more representation in the latent space. This is followed by inverse warping (unwarp) after image translation to restore the original geometry (Figure~\ref{fig:warp_unwarp_pix2pix}). This warp-unwarp strategy is model-agnostic, requires no architectural changes and little additional computation, and can be integrated into any latent image-to-image translation framework. This paper makes the following contributions.



\begin{itemize}
    \item We propose a simple, yet effective, saliency-guided warp–unwarp framework that reallocates spatial representation to salient regions before encoding, counteracting latent spatial shrinkage without enlarging the latent space.
    \item Our approach is model-agnostic, requires no architectural modification, and is compatible with existing latent image-to-image translation frameworks.
    \item Our warping introduces negligible computational overhead (3.0 ms per image) without any additional learnable parameters.
    \item To produce high-quality paired data for human and driving scene relighting, we propose a simple and efficient synthetic data generation pipeline. 
\end{itemize}

Extensive experiments demonstrate strong performance across multiple image-to-image translation tasks, including human relighting, driving scene relighting, and weather or time-of-day translation. Quantitative evaluation using user studies and standard metrics (FID~\cite{heusel2017gans}, KID~\cite{binkowski2018demystifying}, Clean-FID~\cite{parmar2022aliased}, DINO-Struct~\cite{tumanyan2022splicing}, ArcFace~\cite{deng2018arcface}, masked LPIPS~\cite{zhang2018unreasonable}, CLIP score~\cite{radford2021learning,hessel2021clipscore}) shows that our warping-based method consistently outperforms recent state-of-the-arts, including pix2pix-Turbo, CycleGAN-Turbo~\cite{parmar2024one}, IC-Light~\cite{zhang2025scaling} and DreamLight~\cite{liu2026dreamlight}. Although primarily designed for static images, our method can also be applied frame-by-frame to video, demonstrating good temporal consistency and stability (See project webpage for video results). Even without warping, our data generation pipeline for creating high-quality paired images could benefit downstream tasks like scene understanding and image rendering. \textit{Additional results are in Supp.}

\section{Related Works}
\label{sec:related}

\subsection{Image Warping}


Non-uniform spatial transformations were originally developed to correct geometric distortions, as studied in Feature-Based Image Metamorphosis~\cite{beier2023feature} and Digital Image Warping~\cite{wolberg1990digital}. With deep learning, Spatial Transformer Networks~\cite{jaderberg2015spatial} enabled end-to-end learned transformations that intentionally distort images to improve performance. Learning to Zoom~\cite{recasens2018learning} showed the benefit of over-sampling salient regions for multi-resolution reasoning, motivating extensions to detection and segmentation. Following this line, Fovea~\cite{thavamani2021fovea} introduces a dataset-level static \& temporal prior, while Two-Plane Prior~\cite{ghosh2023learned} employs an image-level geometric prior to guide warping. More recently, InstanceWarp~\cite{zheng2025instance} leverages an instance-level object prior for domain-adaptive detection and segmentation.


In comparison, we are the first to apply saliency-guided image warping to generative tasks with three key insights. \textit{(a) Model-Agnostic.} Prior discriminative warping methods like~\cite{thavamani2023learning,zheng2025instance} unwarp in 2D feature space, making them incompatible with 1D token sequences from ViT/DiT backbones, whereas our method unwarps in image space, making it model-agnostic; \textit{(b) Compute-Efficient.} Even if 2D feature maps are generated by the encoder, prior methods must retrain the encoder for warp-compatible features, which is expensive, whereas our method requires only LoRA~\cite{hu2022lora} fine-tuning;  \textit{(c) Improved Fine-Grained Details.} Prior approaches construct saliency maps from coarse-grained signals (e.g., scene-level~\cite{thavamani2021fovea,thavamani2023learning,ghosh2023learned} or instance-level priors~\cite{zheng2025instance}), whereas we derive saliency from fine-grained, part-level cues which show better details (See Figure~\ref{fig:teaser}), while remaining compatible with coarser signals when needed.

\subsection{Text-Guided Image Relighting}



 Text-guided relighting leverages diffusion models to control illumination via natural language, bypassing the need for physical lighting representations which are costly and often unavailable in the wild~\cite{einabadi2021deep}. Text2Relight~\cite{cha2025text2relight} frames portrait illumination as a text-conditional generation problem, while IC-Light~\cite{zhang2025scaling} enforces consistent light transport and DreamLight~\cite{liu2026dreamlight} improves global coherence. However, they often struggle to preserve fine-grained details and spatial consistency.




Our approach differs from prior relighting methods in three key aspects. \textit{(a) Lightweight Data Pipeline.} Prior approaches~\cite{cha2025text2relight,zhang2025scaling,liu2026dreamlight} rely on complex, closed-source, and time-consuming synthetic data pipelines, whereas we leverage off-the-shelf text-to-image models to build a simple, open-source, and efficient pipeline that automatically generates high-quality paired data for diverse relighting conditions; \textit{(b) Data-Efficient.} IC-Light~\cite{zhang2025scaling} and DreamLight~\cite{liu2026dreamlight} require over 10M and 1M training images respectively, whereas our pipeline achieves competitive performance using only ~10K images; \textit{(c) Driving Scene Relighting.} Prior methods focus mainly on human or general object relighting, whereas we demonstrate strong performance on the long-tail driving scenario of roadwork.


\subsection{Weather and Time-of-Day Driving Scene Translation}


Transforming weather and time-of-day conditions in driving scenes predominantly relies on unpaired image-to-image (I2I) translation, as acquiring perfectly paired, pixel-aligned images of the same dynamic scene under drastically different conditions (e.g., clear day vs. heavy fog) is impractical. Existing approaches broadly fall into GAN-based~\cite{goodfellow2014generative} and diffusion-based paradigms~\cite{ho2020denoising}. GAN-based methods address domain discrepancies through cycle consistency (DualGAN~\cite{yi2017dualgan}, CycleGAN~\cite{zhu2017unpaired}), shared  latent spaces (UNIT~\cite{liu2017unsupervised}, MUNIT~\cite{huang2018multimodal}), and contrastive objectives (CUT~\cite{park2020contrastive}, MoNCE~\cite{zhan2022modulated}, AS-IntroVAE~\cite{changjie2023introvae}, TPSeNCE~\cite{zheng2023tpsence}), whereas diffusion-based frameworks (Unit-DDPM~\cite{sasaki2021unit}, DDIBs~\cite{su2022dual}, img2img-turbo~\cite{parmar2024one}) have improved photorealism and stable optimization. Unlike many existing I2I methods that struggle to preserve foreground structures under weather and time-of-day transformations, we leverage image warping to maintain structural consistency while enabling realistic scene-level changes.

\begin{figure}[t]
    \centering
    \includegraphics[width=\linewidth]{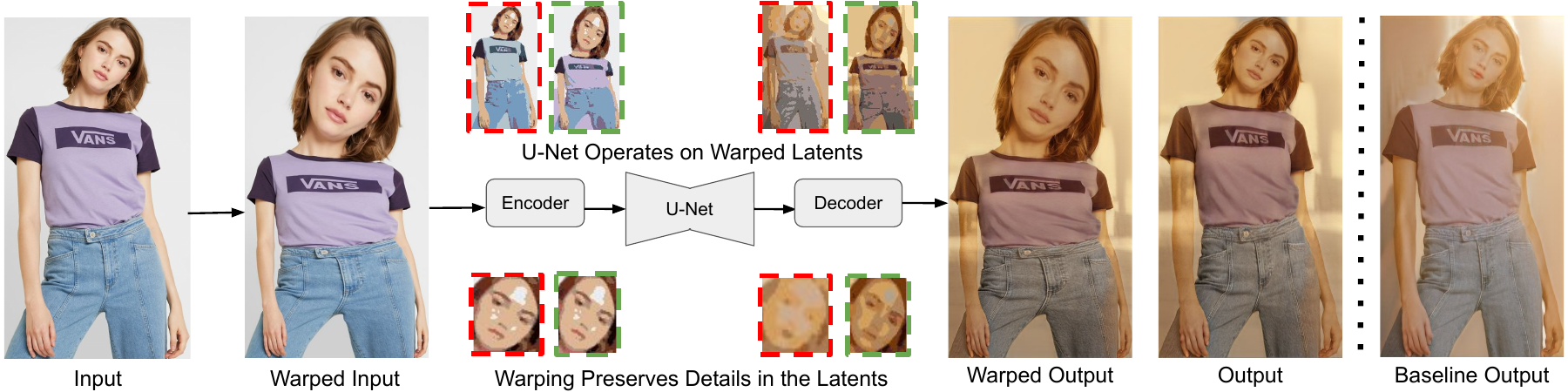}
    \vspace{-0.5cm}
    \caption{\textbf{Overview of our warp-unwarp approach for latent diffusion-based image-to-image translation}. Latent diffusion models encode the input into a spatially downsampled latent space (typically 8×), leaving small but detail-rich salient regions like faces with too few latent pixels to reconstruct details like eyes and expressions. We apply a content-aware warp before encoding to enlarge salient regions, and an inverse warp after translation to restore the original geometry. This warp-unwarp strategy is model-agnostic, requires no architectural changes, very little additional computation, and can be integrated into any latent  image-to-image translation framework. Original latents in \textcolor{red}{\textbf{red}}, and warped latents in \textcolor{warpgreen}{\textbf{green}}. Best viewed when zoomed in.}
   \label{fig:warp_unwarp_pix2pix}
\end{figure}

\section{Method}
\label{sec:method}

\subsection{Overview}
\label{subsec:method_overview}

Figure~\ref{fig:warp_unwarp_pix2pix} illustrates the motivation and architecture of our warp-unwarp framework, which addresses the loss of fine details caused by spatial downsampling in latent diffusion-based image translation by enlarging salient regions before encoding and restoring the original geometry after translation.

In Section~\ref{subsec:synthetic_pairs}, we first construct high-quality synthetic paired data for human relighting and driving scene relighting using a unified FLUX-based pipeline. In Section~\ref{subsec:saliency}, we define a part-level saliency formulation that identifies semantically important regions. In Section~\ref{subsec:warping}, we describe the image warping and unwarping mechanism driven by this saliency map. Finally, in Section~\ref{subsec:integration}, we show how the proposed warp–relight–unwarp pipeline integrates seamlessly into standard image-to-image translation frameworks.


\subsection{Synthetic Paired Data Generation}
\label{subsec:synthetic_pairs}

Real paired relighting data is difficult to obtain, as capturing the same scene under multiple controlled lighting conditions while preserving geometry and foreground consistency is labor-intensive and often infeasible. Consequently, recent relighting approaches such as IC-Light~\cite{zhang2025scaling}, Text2Relight~\cite{cha2025text2relight}, and DreamLight~\cite{liu2026dreamlight} rely on synthetic paired datasets generated through complex and computationally intensive pipelines that are time-consuming and resource-heavy. 

\begin{figure}[t]
    \centering
        \includegraphics[width=\linewidth]{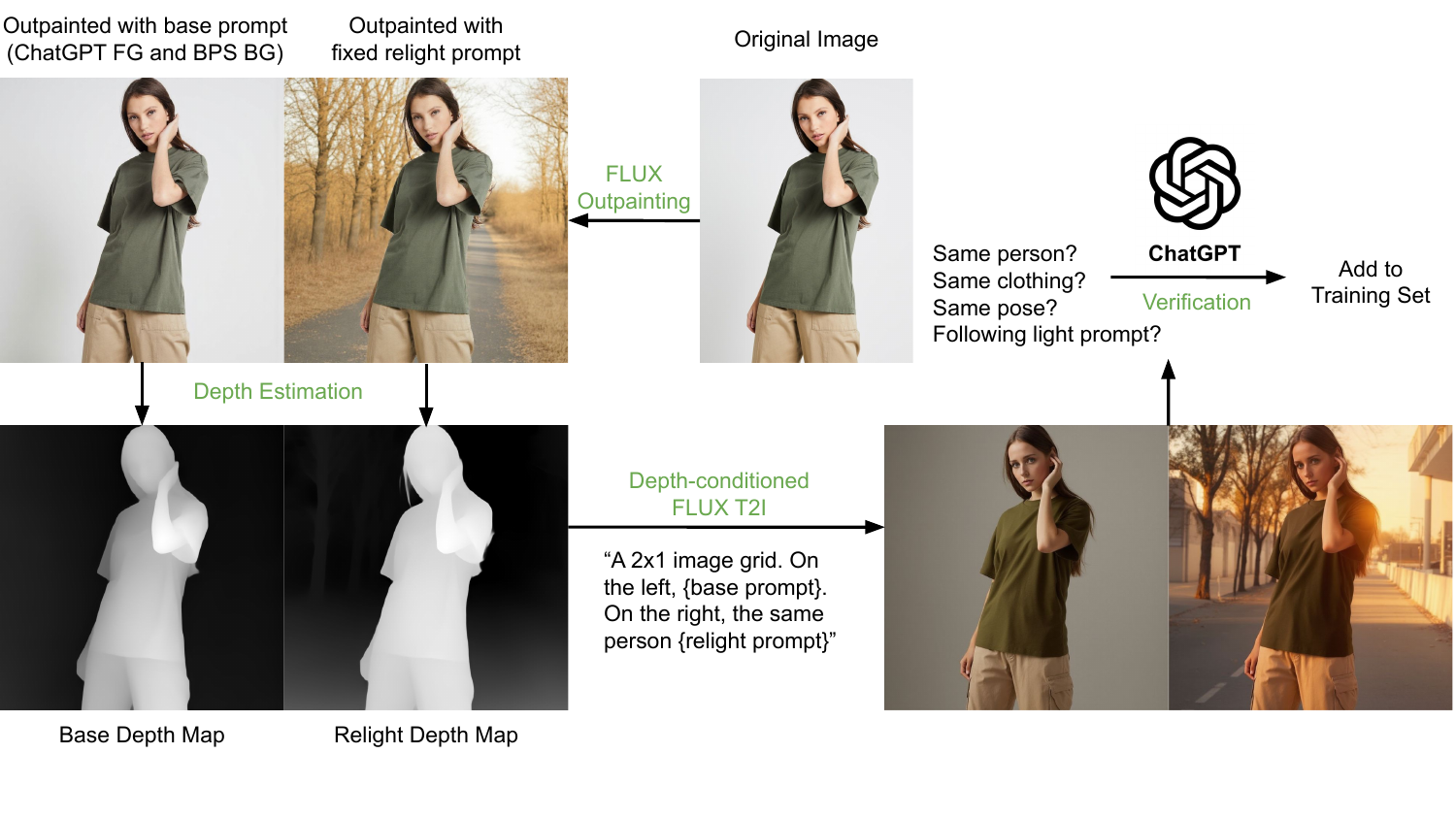}
     \vspace{-1.0cm}
    \caption{\textbf{Our pipeline for generating synthetic paired training images.} We first perform FLUX outpainting to expand the scene while GPT provides text annotations that explicitly separate foreground and background descriptions. We then apply Background Prompt Substitution (BPS) to replace the annotated background with a neutral, descriptive prompt. Next, we run Depth Anything on the outpainted images to obtain depth maps, and feed them into FLUX depth-conditioned generation to synthesize a 2×1 image pair depicting the same person under different backgrounds and lighting. Finally, we use ChatGPT to verify whether the generated pair preserves identity, pose, and prompt fidelity, and keep only approved pairs for training. }
    \label{fig:flux_pair_generation}
\end{figure}

To address these limitations, we develop a simple and efficient pipeline to generate high-quality synthetic data for training paired I2I models such as pix2pix-Turbo~\cite{parmar2024one} under two relighting scenarios: human relighting and driving scene relighting. Both scenarios are built upon a unified FLUX-based~\cite{labs2025flux} generation framework. The pipeline illustrated in Figure~\ref{fig:flux_pair_generation} is designed for human relighting, where we aim to preserve the same foreground while generating a different background under new lighting conditions. In contrast, for driving scene relighting, we remove the outpainting and depth estimation steps and directly apply FLUX text-to-image generation, producing images with the same foreground and background but different lighting. 

\begin{figure}[ht]
    \centering
    \includegraphics[width=0.95\linewidth]
    {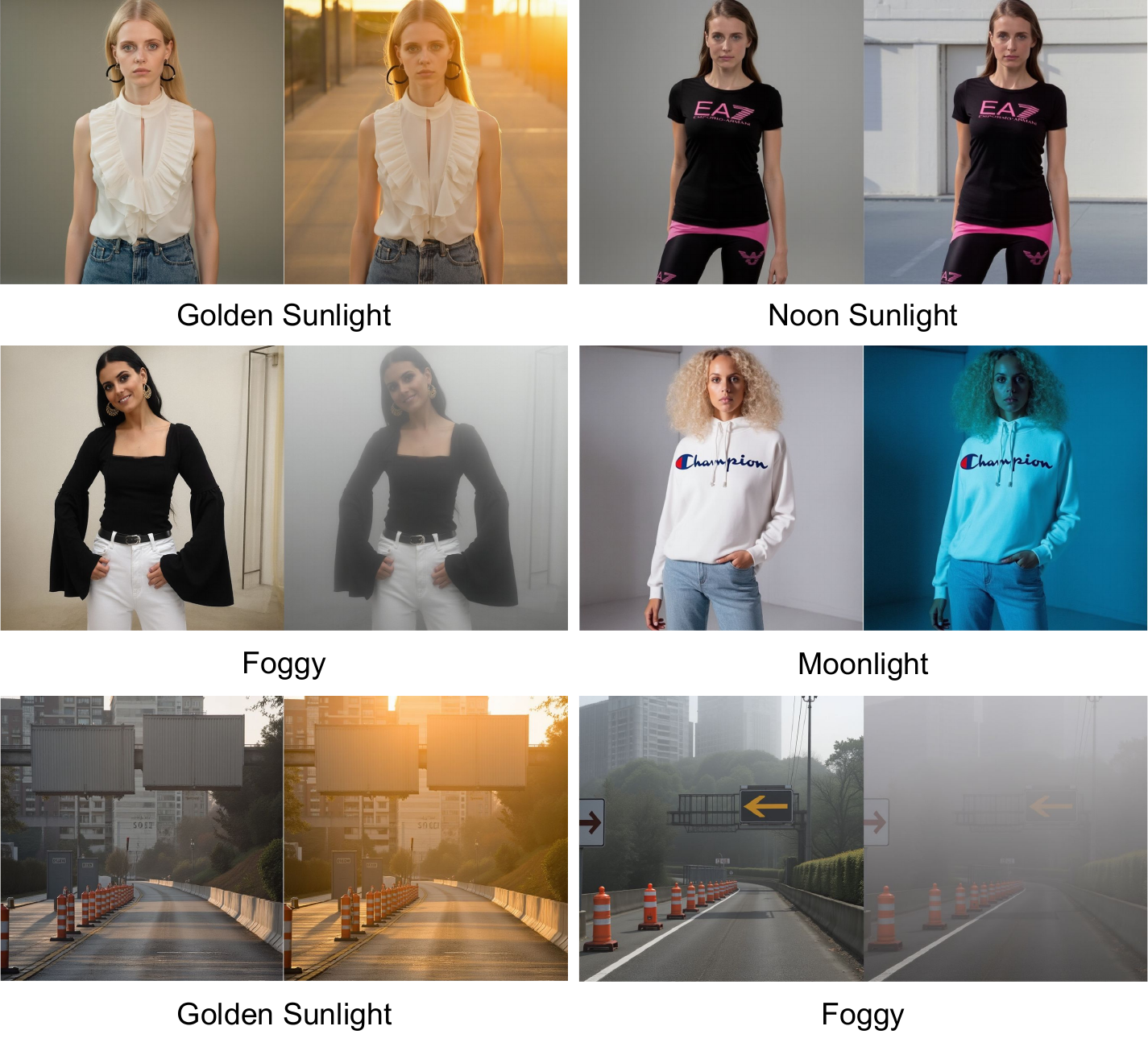}
    \vspace{-0.3cm}
    \caption{\textbf{Synthetic images generated from our pipeline}. Our pipeline produces high-quality training pairs for both human (top two rows) and driving scenes (bottom row). For each pair, the left image is a synthetic base image and the right image is the synthetically relit result under the target lighting condition. Our image pairs preserve foreground identity, maintain strong visual quality, and follow the lighting prompts. For driving scenes, we keep the background consistent without outpainting, while human images allow background outpainting for more flexible appearance. }
    \label{fig:flux_gen_images}
\end{figure}

Empirically, the paired I2I model achieves robust relighting performance using only 10K synthetic training pairs. These pairs can be generated in approximately one day using eight NVIDIA RTX A6000 GPUs. Example synthetic images generated from our pipeline are shown in Figure~\ref{fig:flux_gen_images}.



\noindent \textbf{Image Outpainting via FLUX.}
We use outpainting to generate relit variants for controlled lighting changes while keeping the original foreground unchanged. Specifically, we apply \texttt{FLUX.1-Fill-dev} on an expanded canvas with a foreground mask. A base prompt is derived from a ChatGPT description of each image using a structured template that specifies the foreground person’s attributes (pose, clothing, accessories, hair, and expression) and the background scene description, and a fixed relighting prompt is used to produce the relit background for each lighting condition.

\noindent \textbf{Background Prompt Substitution (BPS).}
The VITON-HD~\cite{choi2021viton} dataset contains images with a plain white background. When captions are generated automatically from ChatGPT, this background is typically described as \textit{“empty background with white studio light.”} Consequently, the generated base images also contain white background. Training an I2I model to translate such images into relit outputs with diverse backgrounds creates an ill-posed one-to-many mapping, which can destabilize training. To mitigate this issue, we propose Background Prompt Substitution (BPS), which replaces this simple background description with a fixed but neutral and more descriptive background prompt: \textit{“background of a neutral indoor studio with visible walls and a flat floor plane, softly lit with even ambient light.”}

\noindent \textbf{Depth Estimation via Depth Anything.}
We compute depth maps for both the outpainted base and relit image using the Depth Anything-Large model~\cite{yang2024depth}. These depth maps provide geometric conditioning for the subsequent generation step while maintaining consistent foreground structure across lighting variants.

\noindent \textbf{Depth-Conditioned Image Generation via FLUX.}
Although the outpainted images share the same foreground and different backgrounds, they are not ideal training pairs because the foreground in the relit image is not explicitly relit by the relighting prompt. To address this, we concatenate the base and relit depth maps into a 2×1 grid (left–right) and use this combined depth map as the conditioning input. We pair it with a structured prompt (\textit{“A 2×1 image grid; on the left, {base prompt}; on the right, the same person {relight prompt}”}) and generate the final image using \texttt{FLUX.1-Depth-dev}, preserving identity, clothing, and pose while producing globally different lighting across the two halves.

\noindent \textbf{ChatGPT Image Filtering.}
Not all generated images are high-quality enough to serve as pseudo ground truth for training, and manually filtering low-quality samples at scale would be impractical. Therefore, we use ChatGPT to automatically filter the generated base images and relit images by verifying lighting consistency with the target description and checking identity, clothing, and pose consistency. Samples are retained only when all criteria are satisfied.

\subsection{Part-Level Saliency Guidance}
\label{subsec:saliency}

Higher saliency corresponds to stronger spatial warping, resulting in greater magnification of the selected region. Therefore, the saliency map should be derived from semantically important cues and tailored to the target task. For human-centric relighting, we assign higher saliency to fine-grained, part-level regions such as faces, especially the eyes. For driving scenes, saliency focuses on object-level regions rather than small part-level details.

The static prior~\cite{thavamani2021fovea} relies on a single dataset-level saliency map for all images, causing warping to incorrectly emphasize average object locations when a specific image deviates from this distribution, while the geometric prior~\cite{ghosh2023learned} depends on vanishing point cues, not applicable in portrait or non-perspective scenes and tends to oversample distant regions that are not necessarily important. 

Our approach is most similar to the instance prior in Instance-Warp~\cite{zheng2025instance}, but extends saliency guidance to fine-grained parts (e.g., eyes and faces) rather than being limited to whole objects. Unlike Instance-Warp, we do not rely on dataset statistics (e.g., object size distributions), as we found that a simple saliency bandwidth of 128 is sufficient for effective warping and unwarping.

Given an image, we aim to oversample important parts like faces, but inside the faces, we want to oversample \textit{more} for \textit{very} important parts like eyes.  Following prior work~\cite{thavamani2021fovea, ghosh2023learned, zheng2025instance}, we use kernel density estimation to model the bounding box saliency as,

\begin{equation}
S = \sum_{(c_i, w_i, h_i) \in y_a} 
N \left( 
c_i, 
b \begin{bmatrix} 
w_i & 0 \\ 
0 & h_i 
\end{bmatrix} 
\right)
\end{equation}

Where $c_i$, $w_i$, and $h_i$ represent the center, width, and height of the bounding box, $N(\mu, \Sigma)$ is the normal distribution, and $b$ is the warping bandwidth. Intuitively, larger bandwidth leads to less intense warping.

\begin{figure}[t]
    \centering
    \includegraphics[width=\linewidth]{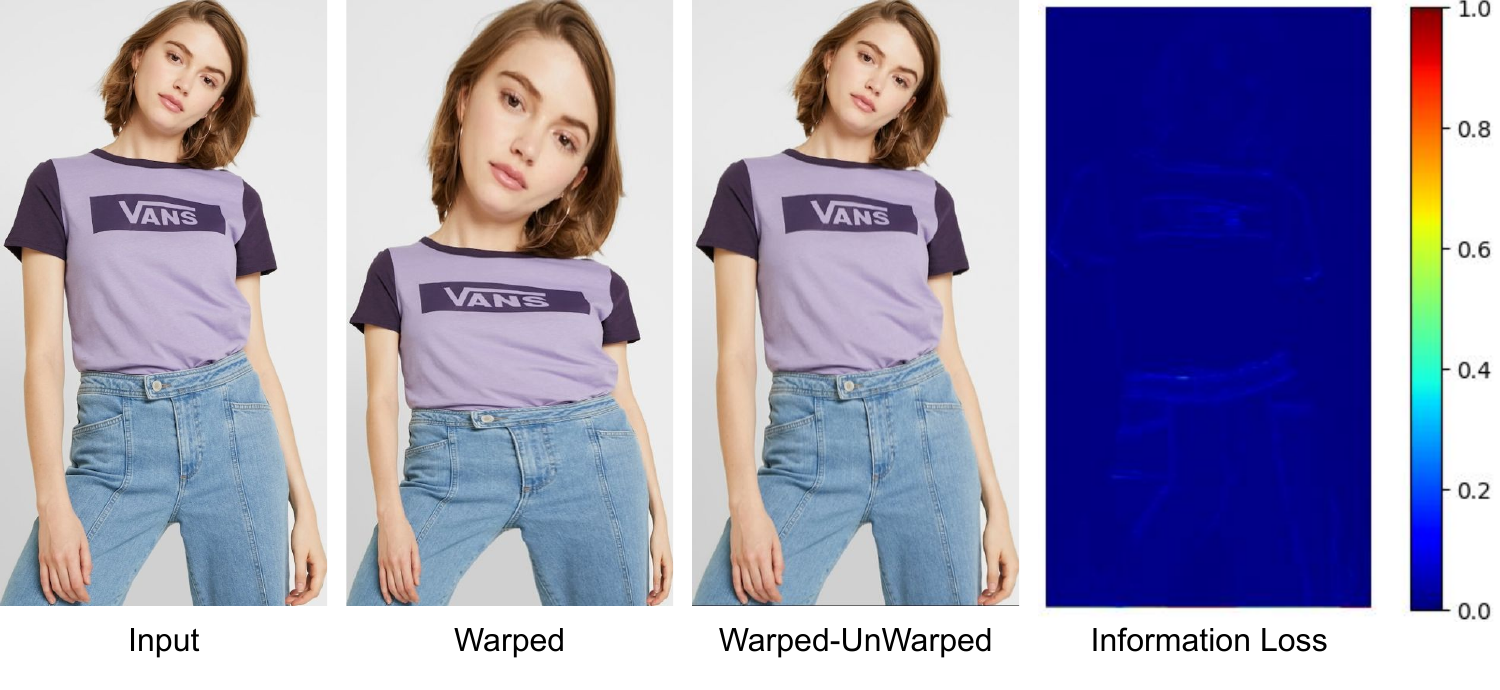}
    \vspace{-0.7cm}
    \caption{\textbf{Warping followed by unwarping introduces minimal information loss.} From left to right, we show the original cropped image, the warped image, the reconstructed image after unwarping, and the corresponding pixel-wise difference heatmap, which indicates minimal information loss.}
    \label{fig:info_loss}
\end{figure}

\subsection{Image Warping and Unwarping}
\label{subsec:warping}

\noindent \textbf{Image Warping.} Let $I$ denote the input image and $S$ denote the saliency map defined over the output resolution. We parameterize the warp through an inverse mapping ($T_S^{-1}$) and obtain the warped image by backward resampling,

\begin{equation}
I'(u) = W_T(I) = I\!\left(T_S^{-1}(u)\right)
\label{eq:warp}
\end{equation}

where $u$ indexes pixel locations. Image warping is efficient, since it introduces an additional 3.0 ms latency per image and has no additional learned parameters, which is negligible for most image translation applications.

\noindent \textbf{Image Unwarping.} I2I Translation demands pixel-level fidelity. Unlike prior warping approaches that operate in feature space (e.g., LZU~\cite{thavamani2023learning}, Instance-Warp~\cite{zheng2025instance}) or prediction space (e.g., Fovea~\cite{thavamani2021fovea}, TPP~\cite{ghosh2023learned}), we perform unwarping directly in image space to better preserve fine-grained details. 

Since the inverse warp $T^{-1}$ has no closed-form solution, we follow LZU~\cite{thavamani2023learning} and approximate it with $\tilde{T}^{-1}$, a piecewise tiling of locally invertible bilinear mappings, ensuring that both forward and inverse transformations exist. 

\begin{equation}
T(u)  \approx \tilde{T}(u) = 
\begin{cases} 
\tilde{T}^{-1}_{ij}(u), & \text{if } u \in \text{Range}(\tilde{T}_{ij}) \\
0, & \text{otherwise}
\end{cases}
\end{equation}

The final image is obtained via backward resampling using $\tilde{T}^{-1}$, effectively reversing the warping operation. Our insight is that this unwarping operation, while being an approximation, is reasonable for inversion while losing little fidelity (See Figure~\ref{fig:info_loss}).  Moreover, as the inversion operation is differentiable, the diffusion model learns to compensate for any artifacts caused by warping and unwarping. Finally, unwarping remains efficient, introducing a negligible 3.0 ms latency per image and has no learned parameters.

\begin{table}[t]
\centering
\setlength{\tabcolsep}{6pt}
\caption{\textbf{Task-specific configurations of our warping framework.} For different tasks, we adopt different base I2I models, saliency regions, and saliency signals. The saliency signal is used to derive a saliency map over the specified region, which guides image warping and unwarping. When available, ground-truth annotations are used as the saliency signal; otherwise, detector-generated bounding boxes are employed.}
\label{tab:warp_configs}
\resizebox{\linewidth}{!}{
\begin{tabular}{lllll}
\toprule
\textbf{Task} & \textbf{Type} & \textbf{Base I2I Model} & \textbf{Saliency Region} & \textbf{Saliency Signal} \\
\midrule
Human Relight & Paired & pix2pix-Turbo & Face \& Eyes & InsightFace (buffalo\_l) \\
Driving Relight & Paired & pix2pix-Turbo & Object & YOLO-World (yolov8x)  \\ 
Weather/ToD Driving & Unpaired & CycleGAN-Turbo & Object & GT bboxes \\
\bottomrule
\end{tabular}
}
\end{table}

\subsection{Adding Warping into Image-to-Image Translation}
\label{subsec:integration}

Our warping framework is model-agnostic, and integrates seamlessly into both paired and unpaired image-to-image translation settings. As summarized in Table~\ref{tab:warp_configs}, the only \textit{task-specific} components are the base I2I model and the choice of saliency region and signal. All other steps, including saliency map construction, and the warp–relight–unwarp pipeline, remain identical across tasks. Letting the task dictate which regions to warp is often desirable, since it focuses detail where it matters most for that specific problem (e.g., face for human, car for driving).

\begin{wrapfigure}{r}{0.25\linewidth}
    \vspace{-\baselineskip}
    \vspace{-\baselineskip}
    \centering
    \hspace{-0.1in}\includegraphics[width=\linewidth]{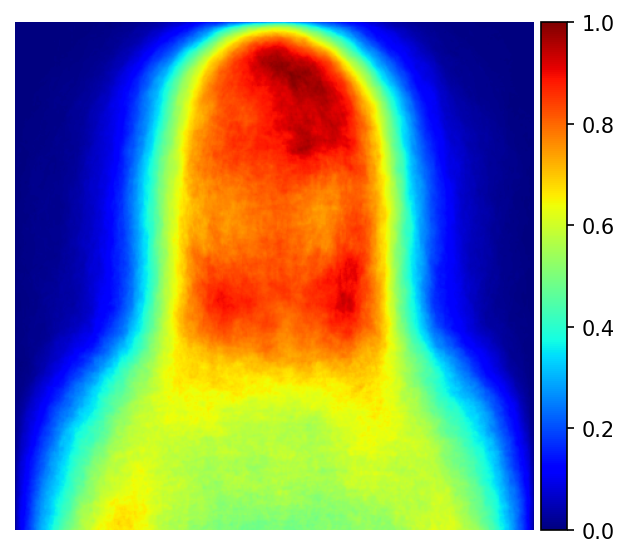}
    \vspace{-0.2in}
\end{wrapfigure} 

That said, task-specific saliency isn't required. For a \textit{task-independent} approach, choices are: \textit{(a)} task-independent detector, where bounding-box size relative to image size (e.g., 10\%) is used to flag small salient regions;
\textit{(b)} pixel-wise error (input - neutral-prompt output) indicates regions for improvement. For human photos, the face region has high error (right) and can guide saliency.

\section{Experiments}
\label{sec:experiment}

\subsection{Human Relighting}

\noindent \textbf{Datasets.} We evaluate human relighting on the VITON-HD~\cite{choi2021viton} and StreetTryOn~\cite{cui2025street} datasets. For dataset preparation, we first annotate textual relighting prompts using ChatGPT (annotation format in Supp.) and generate foreground masks with Grounded-SAM2~\cite{ren2024grounded}. Using these annotations, we construct synthetic paired data by running our text-to-image pipeline on VITON-HD.

\noindent \textbf{Training and Testing.} We train pix2pix-Turbo~\cite{parmar2024one} on synthetic pairs with our warp–relight–unwarp framework, where saliency is computed using detected bounding boxes of the face and eyes. For testing, we evaluate on real images from the VITON-HD test split and the StreetTryOn~\cite{cui2025street} test set with no fine-tuning.

\noindent \textbf{Evaluation Metrics.} We conduct a user study with 60 participants with undergraduate computer vision background, as evaluating technical aspects such as semantic consistency and lighting faithfulness requires domain knowledge. Participants rate results on a 1–5 scale (worst to best) across four criteria: 1) person identity preservation, 2) clothing identity preservation, 3) overall image quality, and 4) lighting faithfulness (i.e., alignment between lighting and text prompt). Additional quantitative metrics such as ArcFace identity similarity~\cite{deng2018arcface}, masked LPIPS~\cite{zhang2018unreasonable}, CLIP score~\cite{radford2021learning,hessel2021clipscore}, and GPT-4o evaluation are in Supp.

\begin{figure}[p]
\centering
\includegraphics[width=0.94\linewidth]{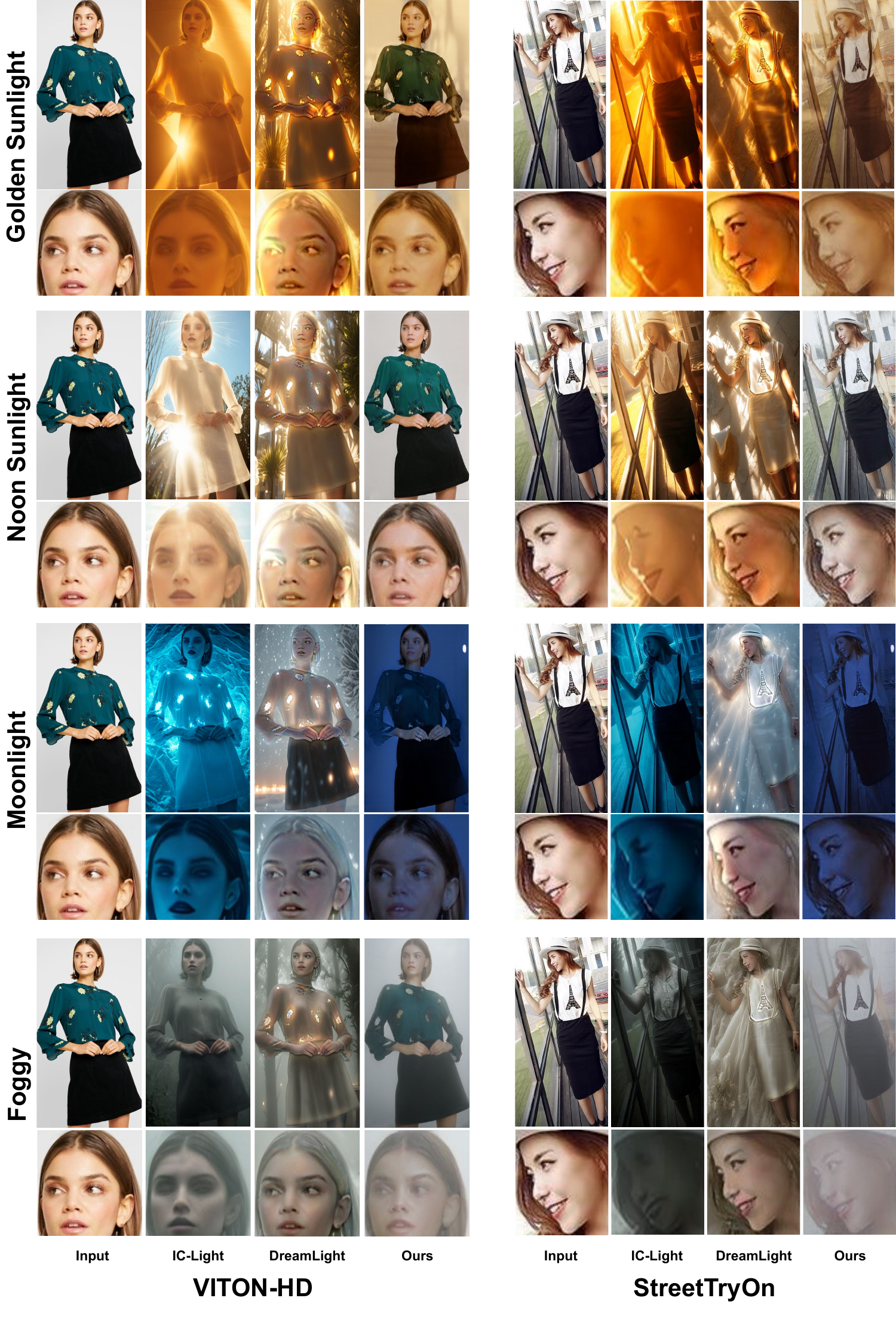}
\vspace{-0.7cm}
\caption{\textbf{Qualitative Comparison on Human Relighting.} We show one example from VITON-HD~\cite{choi2021viton} and another from StreetTryOn~\cite{cui2025street} under different lighting prompts. IC-Light~\cite{zhang2025scaling} and DreamLight~\cite{liu2026dreamlight} fail to preserve clothing identity and exhibit poor lighting faithfulness to the prompts. IC-Light also alters person identity due to distorted facial features. Our method better preserves identity and lighting fidelity, resulting in higher image quality. Please see detailed relighting prompts in Supp.}
\label{fig:vis_viton_street}
\end{figure}

\noindent \textbf{Results.} Qualitative comparisons are shown in Figure \ref{fig:vis_viton_street}, and quantitative results are summarized in Table \ref{tab:human_relighting_user_study}. Our method achieves consistently strong performance across all four criteria, with notably better person and cloth identity preservation, higher image quality, and stronger lighting faithfulness to the relit prompt, outperforming IC-Light~\cite{zhang2025scaling} and DreamLight~\cite{liu2026dreamlight} by a clear margin.

\begin{table}[t]
\caption{\textbf{User study results on human relighting.} Our method utilize a pix2pix-Turbo baseline trained with synthetic pairs generated by our pipeline and evaluated on VITON-HD~\cite{choi2021viton} and StreetTryOn~\cite{cui2025street} datasets. Scores are averaged over four lighting prompts: golden sunlight, noon sunlight, moonlight, and foggy conditions. The proposed warping and Background Prompt Substitution (BPS) improve performance. }
\resizebox{\linewidth}{!}{
\begin{tabular}{lcccccccccc}
\toprule
& \multicolumn{5}{c}{\textbf{VITON-HD}}
& \multicolumn{5}{c}{\textbf{StreetTryOn}} \\
\cmidrule(lr){2-6}\cmidrule(lr){7-11}
Methods
& \begin{tabular}[c]{@{}c@{}}Person\\Identity$\uparrow$\end{tabular}
& \begin{tabular}[c]{@{}c@{}}Cloth\\Identity$\uparrow$\end{tabular}
& \begin{tabular}[c]{@{}c@{}}Image\\Quality$\uparrow$\end{tabular}
& \begin{tabular}[c]{@{}c@{}}Lighting\\Faithfulness$\uparrow$\end{tabular}
& Avg.
& \begin{tabular}[c]{@{}c@{}}Person\\Identity$\uparrow$\end{tabular}
& \begin{tabular}[c]{@{}c@{}}Cloth\\Identity$\uparrow$\end{tabular}
& \begin{tabular}[c]{@{}c@{}}Image\\Quality$\uparrow$\end{tabular}
& \begin{tabular}[c]{@{}c@{}}Lighting\\Faithfulness$\uparrow$\end{tabular}
& Avg. \\
\midrule
IC-Light~\cite{zhang2025scaling}
& 3.43 & 3.46 & 3.47 & 3.51 & 3.47
& 3.33 & 3.20 & 3.25 & 3.33 & 3.28 \\

DreamLight~\cite{liu2026dreamlight}
& 3.52 & 3.50 & 3.50 & 3.41 & 3.48
& 3.16 & 3.13 & 3.07 & 3.12 & 3.12 \\

\midrule

Ours (No BPS)
& 3.70 & 3.72 & 3.62 & 3.64 & 3.67
& 3.60 & 3.59 & 3.49 & 3.53 & 3.55 \\

Ours (No Warp)
& 3.60 & 3.62 & 3.48 & 3.54 & 3.56
& 3.60 & 3.63 & 3.49 & 3.55 & 3.57 \\

Ours
& \textbf{4.36} & \textbf{4.43} & \textbf{4.31} & \textbf{4.21} & \textbf{4.33}
& \textbf{4.31} & \textbf{4.29} & \textbf{4.18} & \textbf{4.14} & \textbf{4.23} \\
\bottomrule
\end{tabular}
}
\label{tab:human_relighting_user_study}
\end{table}

\subsection{Ablation Studies}

We conduct ablations to demonstrate the effectiveness of our warping and Background Prompt Substitution (BPS). User study scores are reported in Table \ref{tab:human_relighting_user_study}, showing that both BPS and warping improve image relighting. Additional ablations are in Supp.: \textit{(a)} Our method's robustness to detection error and bandwidth values; \textit{(b)} pix2pix-Turbo with our warping vs. ChatGPT image-to-image.



\subsection{Inference Latency} 

As shown in Table~\ref{tab:inf_latency}, our proposed warp-unwarp module introduces a negligible additional latency of only 0.006\,s (3\,ms warping, 3\,ms unwarping), making it a promising plug-and-play module suitable for real-time applications. 

\begin{table}[t]
\setlength{\tabcolsep}{4pt}
\centering
\caption{\textbf{Inference latency for $768 \times 1024$ resolution images (VITON dataset).} }
\label{tab:latency_comparison}
\resizebox{\linewidth}{!}{
\begin{tabular}{lccccc} 
\toprule
\textbf{Model} & \begin{tabular}{@{}c@{}}ChatGPT \\ img2img\end{tabular} & IC-Light & DreamLight & pix2pix-Turbo & \textbf{\begin{tabular}{@{}c@{}}pix2pix-Turbo \\ + Warp-Unwarp\end{tabular}} \\ 
\midrule
\textbf{Latency (s)} & $\sim$30.0 & 1.237 & 1.954 & 0.892 & \textbf{\begin{tabular}{@{}c@{}}0.898 \\ (+0.006)\end{tabular}} \\ 
\bottomrule
\end{tabular}
}
\label{tab:inf_latency}
\end{table}

\subsection{Weather and Time-of-Day Driving Scene Translation}

\noindent \textbf{Datasets.} We evaluate on BDD100K~\cite{yu2020bdd100k} (clear day, clear night, rainy day), Cityscapes~\cite{cordts2016cityscapes}, DarkZurich~\cite{sakaridis2019guided}, ACDC~\cite{sakaridis2021acdc} (foggy), and Dense Fog~\cite{bijelic2020seeing}. 

\noindent \textbf{Training and Testing.} We train CycleGAN-Turbo~\cite{parmar2024one} with our warp–unwarp framework, where saliency is computed using ground-truth bounding boxes of foreground objects. Models are evaluated on each dataset’s official test split.


\noindent \textbf{Evaluation Metrics.} We report FID~\cite{heusel2017gans}, KID~\cite{binkowski2018demystifying}, Clean-FID~\cite{parmar2022aliased}, and DINO-Struct~\cite{tumanyan2022splicing} to measure both visual fidelity and structural consistency.



\noindent \textbf{Results.} Quantitative results are summarized in Table~\ref{tab:intra_dataset_translation} for intra-dataset experiments and Table~\ref{tab:cross_dataset_translation} for cross-dataset experiments. Our warping consistently improves FID, KID, and Clean-FID, indicating better alignment with real images in the target domain. DINO-Struct also improves in most cases, indicating better preservation of structural information from the source image. The only exception is DarkZurich $\rightarrow$ Cityscapes, where the metric is less reliable due to the reduced robustness of DINO features under low-light conditions~\cite{zhang2025wakeup}. Finally, replacing GT bboxes with YOLO-World~\cite{cheng2024yolo} detected bboxes yields comparable gains, showing our method does not rely on ground-truth annotations. Qualitative results are in Supp.

\begin{table}[t]
\setlength{\tabcolsep}{4pt}
\caption{\textbf{Quantitative results on weather and time-of-day driving scene translation (intra-dataset).} KID is reported $\times$1000 and DINO-Struct $\times$100.}
\resizebox{\linewidth}{!}{
\begin{tabular}{lcccccccccccc}
\toprule
& \multicolumn{4}{c}{\textbf{\begin{tabular}[c]{@{}c@{}}BDD100K\\(Day $\rightarrow$ Night)\end{tabular}}}
& \multicolumn{4}{c}{\textbf{\begin{tabular}[c]{@{}c@{}}BDD100K\\(Night $\rightarrow$ Day)\end{tabular}}}
& \multicolumn{4}{c}{\textbf{\begin{tabular}[c]{@{}c@{}}BDD100K\\(Clear $\rightarrow$ Rainy)\end{tabular}}} \\
\cmidrule(lr){2-5}\cmidrule(lr){6-9}\cmidrule(lr){10-13}
Methods
& FID$\downarrow$ & KID$\downarrow$ & \begin{tabular}[c]{@{}c@{}}Clean-\\FID$\downarrow$\end{tabular} & \begin{tabular}[c]{@{}c@{}}DINO\\Struct.$\downarrow$\end{tabular}
& FID$\downarrow$ & KID$\downarrow$ & \begin{tabular}[c]{@{}c@{}}Clean-\\FID$\downarrow$\end{tabular} & \begin{tabular}[c]{@{}c@{}}DINO\\Struct.$\downarrow$\end{tabular}
& FID$\downarrow$ & KID$\downarrow$ & \begin{tabular}[c]{@{}c@{}}Clean-\\FID$\downarrow$\end{tabular} & \begin{tabular}[c]{@{}c@{}}DINO\\Struct.$\downarrow$\end{tabular} \\
\midrule
CycleGAN-Turbo~\cite{parmar2024one}
& 19.2 & 8.08 & 31.3 & 3.00
& 41.5 & 27.7 & 45.2 & 3.80
& 57.7 & 9.47 & 57.2 & 1.50 \\
\midrule
+Warp Det bbox
& 17.7 & 6.70 & 17.3 & 2.95
& 36.8 & 24.1 & 35.6 & 3.55 
& 56.0 & 6.93 & 56.3 & 0.86  \\
+Warp GT bbox
& \textbf{17.5} & \textbf{6.61} & \textbf{17.2} & \textbf{2.92}
& \textbf{36.3} & \textbf{23.8} & \textbf{35.4} & \textbf{3.51}
& \textbf{55.9} & \textbf{6.91} & \textbf{55.9} & \textbf{0.84} \\
\bottomrule
\end{tabular}
}
\label{tab:intra_dataset_translation}
\end{table}

\begin{table}[t]
\setlength{\tabcolsep}{4pt}
\caption{\textbf{Quantitative results on weather and time-of-day driving scene translation (cross-dataset).} KID is reported $\times$1000 and DINO-Struct $\times$100.}
\resizebox{\linewidth}{!}{
\begin{tabular}{lcccccccccccc}
\toprule
 & \multicolumn{4}{c}{\textbf{Cityscapes $\rightarrow$ DarkZurich}}
 & \multicolumn{4}{c}{\textbf{DarkZurich $\rightarrow$ Cityscapes}}
 & \multicolumn{4}{c}{\textbf{Cityscapes $\rightarrow$ ACDC Foggy}} \\
\cmidrule(lr){2-5}\cmidrule(lr){6-9}\cmidrule(lr){10-13}
Methods
& FID$\downarrow$ & KID$\downarrow$ & \begin{tabular}[c]{@{}c@{}}Clean-\\FID$\downarrow$\end{tabular} & \begin{tabular}[c]{@{}c@{}}DINO\\Struct.$\downarrow$\end{tabular}
& FID$\downarrow$ & KID$\downarrow$ & \begin{tabular}[c]{@{}c@{}}Clean-\\FID$\downarrow$\end{tabular} & \begin{tabular}[c]{@{}c@{}}DINO\\Struct.$\downarrow$\end{tabular}
& FID$\downarrow$ & KID$\downarrow$ & \begin{tabular}[c]{@{}c@{}}Clean-\\FID$\downarrow$\end{tabular} & \begin{tabular}[c]{@{}c@{}}DINO\\Struct.$\downarrow$\end{tabular} \\
\midrule
CycleGAN-Turbo~\cite{parmar2024one}
& 159.5 & 98.1 & 160.3 & 4.35
& 221.4 & 157.8 & 209.6 & \textbf{2.22}
& 308.8 & 340.9 & 286.5 & 9.18 \\
\midrule
+Warp Det bbox
& 154.1 & 93.6 & 155.0 & 3.62
& 213.7 & 146.6 & 206.5 & 2.99
& 182.7 & 108.7 & 169.5 & 3.72 \\
+Warp GT bbox
& \textbf{153.3} & \textbf{93.2} & \textbf{153.4} & \textbf{3.20}
& \textbf{212.6} & \textbf{144.9} & \textbf{204.6} & 2.75
& \textbf{169.2} & \textbf{96.6} & \textbf{154.4} & \textbf{2.53} \\
\bottomrule
\end{tabular}
}
\label{tab:cross_dataset_translation}
\end{table}

\subsection{Driving Scene Relighting}

\noindent \textbf{Datasets.} We use the ROADWork~\cite{ghosh2025roadwork} dataset due to its rich ground-truth text annotations for generating relighting training pairs. We construct synthetic paired data by running our text-to-image pipeline on ROADWork (Pittsburgh).

\noindent \textbf{Training and Testing.} We train pix2pix-Turbo~\cite{parmar2024one} on synthetic pairs with our warp–relight–unwarp framework, where saliency is computed using YOLO-World~\cite{cheng2024yolo} detected object bounding boxes. Evaluation is at ROADWork (Boston).

\noindent \textbf{Evaluation Metrics.} We follow the same user-study protocol as in human relighting, but replacing person and clothing identity with semantic consistency.

\begin{figure}[p]
\centering
\includegraphics[width=\linewidth]{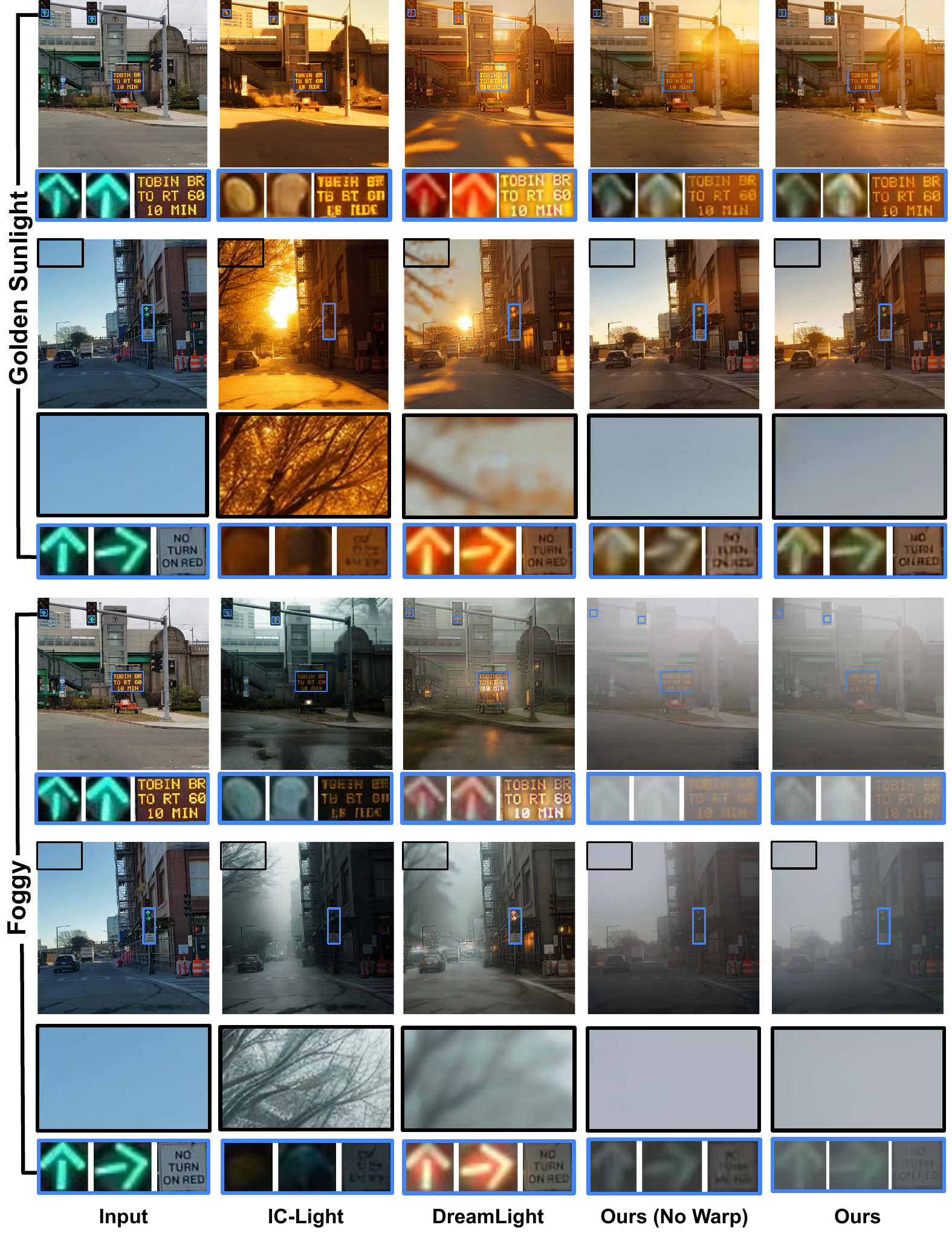}
\vspace{-0.6cm}
\caption{\textbf{Qualitative comparison on driving scene relighting.} IC-Light~\cite{zhang2025scaling} and DreamLight~\cite{liu2026dreamlight} produce lower image quality and less faithful lighting, and often fail to maintain semantic consistency. For example, IC-Light turns green traffic arrows into circular lights or removes them, while DreamLight changes green arrows into red or yellow ones. Our method has better semantic consistency, image quality, and lighting faithfulness. With our warping, small details such as traffic arrows and the `No Turn on Red' sign become more clearly visible.}
\label{fig:vis_roadwork_merged}
\end{figure}

\noindent \textbf{Results.} Qualitative comparisons are shown in Figure~\ref{fig:vis_roadwork_merged}, and user study results are shown in Table~\ref{tab:roadwork_relighting_study}. Our method, especially with warping, shows better semantic consistency, image quality, and lighting faithfulness, compared with the baseline IC-Light~\cite{zhang2025scaling} and DreamLight~\cite{liu2026dreamlight}.

\begin{table}[t]
\caption{\textbf{User study results on driving scene relighting.} Our method utilizes a pix2pix-Turbo baseline trained with synthetic pairs generated by our pipeline and evaluated on the ROADWork (Boston) dataset under two lighting prompts: golden sunlight and foggy. For driving scene relighting, Background Prompt Substitution (BPS) is not applied because we keep the background consistent and therefore do not perform image outpainting when generating the synthetic training pairs. The proposed warping improves performance, particularly on semantic consistency.}
\resizebox{\linewidth}{!}{
\begin{tabular}{lcccccccc}
\toprule
& \multicolumn{4}{c}{\textbf{ROADWork (Golden Sunlight)}}
& \multicolumn{4}{c}{\textbf{ROADWork (Foggy)}} \\
\cmidrule(lr){2-5}\cmidrule(lr){6-9}
Methods
& \begin{tabular}[c]{@{}c@{}}Semantic\\Consistency$\uparrow$\end{tabular}
& \begin{tabular}[c]{@{}c@{}}Image\\Quality$\uparrow$\end{tabular}
& \begin{tabular}[c]{@{}c@{}}Lighting\\Faithfulness$\uparrow$\end{tabular}
& Avg.
& \begin{tabular}[c]{@{}c@{}}Semantic\\Consistency$\uparrow$\end{tabular}
& \begin{tabular}[c]{@{}c@{}}Image\\Quality$\uparrow$\end{tabular}
& \begin{tabular}[c]{@{}c@{}}Lighting\\Faithfulness$\uparrow$\end{tabular}
& Avg. \\
\midrule
IC-Light~\cite{zhang2025scaling}
& 3.63 & 3.65 & 3.82 & 3.70
& 3.82 & 3.87 & 3.92 & 3.77 \\

DreamLight~\cite{liu2026dreamlight}
& 3.97 & 3.90 & 3.95 & 3.94
& 3.85 & 3.87 & 3.90 & 3.91 \\

\midrule

Ours (No Warp)
& 4.17 & 4.30 & 4.28 & 4.25
& 4.18 & 4.28 & 4.37 & 4.26 \\

Ours
& \textbf{4.53} & \textbf{4.55} & \textbf{4.55} & \textbf{4.54}
& \textbf{4.48} & \textbf{4.45} & \textbf{4.48} & \textbf{4.51} \\
\bottomrule
\end{tabular}
}
\label{tab:roadwork_relighting_study}
\end{table}

\section{Conclusion and Limitations}
\label{sec:conclusion}


In this paper, we propose a simple saliency-guided warp–unwarp framework that reallocates spatial representation toward salient regions before encoding, enabling better preservation of structural details without increasing latent resolution. The warped image is processed by the original diffusion model and then mapped back via an inverse warp. In addition, we propose a simple and efficient outpainting-based synthetic data generation pipeline to produce high-quality paired data for image relighting. Our method is model-agnostic, requires no architectural modification, and introduces negligible computational overhead. Experiments on human relighting, driving scene relighting and translation demonstrate improved structural preservation, lighting faithfulness, and overall image quality. Beyond static images, our framework extends naturally to video via frame-by-frame application with good temporal stability.

\noindent \textbf{Limitations.} Although our warping is model-agnostic, we currently evaluate it only on one-step diffusion baselines (e.g., pix2pix-Turbo and CycleGAN-Turbo~\cite{parmar2024one}) due to computational constraints. Future work could explore integrating our approach with larger, multi-step diffusion models. Additionally, while our framework extends to video on a frame-by-frame basis, it may flicker under fast motion, where stronger temporal priors such as optical flow could improve consistency.



\noindent \textbf{Acknowledgments.} This research was supported in parts by SpreeAI (human relighting), General Motors - Research (driving scenes) and by U.S. Department of Transportation Grant 69A3552344811 and 69A3552348316 through the Carnegie Mellon University Safety21 University Transportation Center.

\clearpage


%
%
\bibliographystyle{splncs04}
\bibliography{main}

\clearpage
\renewcommand{\thesection}{\Alph{section}}
\setcounter{section}{0}
\renewcommand{\thefigure}{\arabic{figure}}
\renewcommand{\thetable}{\arabic{table}}


\title{WarpI2I: Image Warping for Image-to-Image Translation \\ \textit{Supplementary Material}} 

\titlerunning{WarpI2I}

\author{Shen Zheng \and
Anurag Ghosh \and
Gaurav Parmar \and
Srinivasa Narasimhan
}

\authorrunning{Zheng et al.}

\institute{Carnegie Mellon University}

\vspace{-0.2cm}
\section*{Overview}

This supplementary material provides additional context on key design decisions (Section~\ref{supsec:faqs}), robustness under detection error and saliency bandwidth variation (Section~\ref{supsec:robustness}), warp vs. no-warp on fine details (Section~\ref{supsec:warp_detail}), comparisons with ChatGPT img2img (Section~\ref{supsec:ours_chatgpt}), details on our annotation and filtering pipeline (Sections~\ref{supsec:chatgpt_anno}--\ref{supsec:chatgpt_filtering}), relighting prompts (Section~\ref{supsec:text_prompts}), additional details for user study (Section~\ref{supsec:user_study_details}), and additional results across human relighting (Section~\ref{supsec:add_res_human}), driving scene relighting (Section~\ref{supsec:add_res_driving}), and weather and time-of-day translation (Section~\ref{supsec:add_res_weather}).



\section{Design Rationale and Extended Discussion}
\label{supsec:faqs}

\noindent\textit{\textbf{Design Rationale}}

\noindent \textbf{Q1: What role does outpainting play in the data pipeline, given that text-to-image generation alone suffices for driving scenes?}

\noindent \textbf{A1:} For complex driving scenes like ROADWork~\cite{ghosh2025roadwork}, both foreground objects and background environments are naturally diverse, so text-to-image generation alone can produce varied training image pairs. However, human relighting datasets such as VITON-HD~\cite{choi2021viton} and StreetTryOn~\cite{cui2025street} contain a single foreground subject with relatively simple backgrounds. Without outpainting, the generated image pairs often share very similar background layouts, limiting training diversity. Our outpainting step, combined with Background Prompt Substitution (BPS), increases background variability and improves training stability and relighting quality, as confirmed by our user study and visual comparisons.

\medskip
\noindent \textbf{Q2: How does the saliency bandwidth affect warping, and how sensitive is the method to this parameter?}

\noindent \textbf{A2}: Our method is not very sensitive to the warping bandwidth and requires minimal hyperparameter tuning. In practice, we observe that bandwidth values between 64 and 256 produce no significant difference in output quality (See Table~\ref{tab:human_relighting_user_study_new}). This is because as long as the warping enlarges salient objects (e.g., a person) or parts (e.g., face and eyes) beyond a certain threshold in the latent space, their feature representations become sufficiently well preserved.

\medskip
\noindent \textbf{Q3: How does the warping behave when saliency signals are imperfect, such as when face or object detectors produce localized errors?}

\noindent \textbf{A3}: Our warping is fairly robust to small inaccuracies in the saliency signal  (See Table~\ref{tab:human_relighting_user_study_new}). If the bounding boxes are slightly misaligned (e.g., shifted by a few pixels), the warped region is still enlarged and the improvement is typically maintained. If a salient region is missed, it is treated as background with little warping, so the method behaves similarly to the baseline.

\medskip
\noindent\textit{\textbf{Generality \& Robustness}}

\noindent \textbf{Q4: Is the warp-unwarp framework general? Can it work with multi-step diffusion models beyond pix2pix-Turbo and CycleGAN-Turbo?}

\noindent \textbf{A4}: Yes. Our warp-unwarp module is model-agnostic and integrates into other multi-step diffusion pipelines in a plug-and-play manner. In this work, we demonstrate results with pix2pix-Turbo and CycleGAN-Turbo~\cite{parmar2024one}, but the approach generalizes to any latent-based image-to-image translation framework. We focus on these two models mainly due to computational constraints.

\medskip
\noindent\textit{\textbf{Visual Fidelity and Computational Overhead}}

\noindent \textbf{Q5: How do warping and BPS affect relighting quality visually?}

\noindent \textbf{A5:} Warping improves detail preservation during image-to-image translation. Visual comparisons of BPS and warping for human relighting are provided in Section~\ref{supsec:add_res_human}, and for driving scene relighting in Section~\ref{supsec:add_res_driving}.





\medskip
\noindent \textbf{Q6: What is the computational cost of the saliency detectors used for generating saliency maps?}

\noindent \textbf{A6}: The saliency detector adds <0.05 seconds per image, which is negligible compared to diffusion model inference (which typically takes 1 second or more).


\medskip
\noindent\textit{\textbf{Comparison with Commercial Models}}

\noindent \textbf{Q7: How does our method compare with commercial models like ChatGPT img2img for image-to-image translation?}

\noindent \textbf{A7}: Commercial models (e.g., ChatGPT img2img) produce realistic edits but are significantly slower (around 30 seconds per image) and may alter person or clothing identity. Our method runs in under one second and better preserves fine-grained image structure. \textit{Visual comparisons are shown in Section~\ref{supsec:ours_chatgpt}.}


\medskip
\noindent\textit{\textbf{Data Pipeline Quality}}

\noindent \textbf{Q8: What is the quality of the ChatGPT-generated annotations?}

\noindent \textbf{A8}: The annotations are generally high quality and follow a fixed prompt template to ensure consistency. \textit{See Section~\ref{supsec:chatgpt_anno} for details.}

\medskip
\noindent \textbf{Q9: What is the quality and yield of the synthetic pairs from the data pipeline?}

\noindent \textbf{A9}: The synthetic pairs are of high quality: approximately 96\% of generated pairs pass the ChatGPT image filtering stage, resulting in a high usable yield and an efficient data generation process. \textit{See Section~\ref{supsec:chatgpt_filtering} for details.}

\medskip
\noindent \textbf{Q10: How accurate is the ChatGPT-based image filtering?}

\noindent \textbf{A10}: The filtering step is highly reliable. In our manual inspection, around 95\% of the images discarded by ChatGPT agree with human judgment.

\medskip
\noindent \textbf{Q11: Which ChatGPT model is used for annotation and filtering?}

\noindent \textbf{A11}: We use the GPT-4o model through the OpenAI API. Although GPT-4o has been replaced by newer models in the web interface, it remains available via the API and provides strong performance for image understanding.




\clearpage
\section{Robustness to Detection Error and Bandwidth} \label{supsec:robustness}

\vspace{-0.1cm}
\begin{table}[ht]
    \centering
    \caption{\textbf{Robustness to detection error and bandwidth (VITON-HD).} Our method is robust under large Gaussian bbox jitter (std=0.2/0.3 of width), and is not much sensitive to saliency bandwidth (Bw) across 64, 128, and 256. }
    \label{tab:human_relighting_user_study_new}
    \resizebox{\linewidth}{!}{
    \begin{tabular}{lcccccc}
    \toprule
    \textbf{Methods}
    & \textbf{User Study}$\uparrow$
    & \textbf{ArcFace}$\uparrow$
    & \textbf{LPIPS}$\downarrow$
    & \textbf{CLIP}$\uparrow$
    & \textbf{GPT-4o}$\uparrow$ \\
    \midrule
    IC-Light & 3.47 & 0.263 & 0.334 & 18.24 & 60.0 \\
    DreamLight & 3.48 & 0.688 & 0.311 & 19.17 & 66.4 \\
    \midrule
    Ours (no BPS) & 3.67 & 0.613 & 0.269 & 17.74 & 84.2 \\
    Ours (no warp) & 3.56 & 0.566 & 0.264 & 17.86 & 86.1 \\
    \midrule
    Ours (Bw=64) & 4.27 & 0.785 & 0.269 & 20.25 & 93.5 \\
    Ours (Bw=256) & 4.29 & 0.781 & 0.265 & 20.31 & 92.4 \\
    \midrule
    Ours (jitter=0.2) & 4.24 & 0.775 & 0.273 & 20.19 & 92.6 \\
    Ours (jitter=0.3) & 4.15 & 0.767 & 0.281 & 20.07 & 90.9 \\
    \midrule
    Ours (jitter=0; Bw=128) & \textbf{4.36} & \textbf{0.783} & \textbf{0.264} & \textbf{20.35} & \textbf{93.6} \\
    \bottomrule
    \end{tabular}
    }
\end{table}



\section{Warp vs.\ No-Warp on Fine Details}
\label{supsec:warp_detail}

\vspace{-0.1cm}
\begin{figure}
    \centering
    \includegraphics[width=\linewidth]{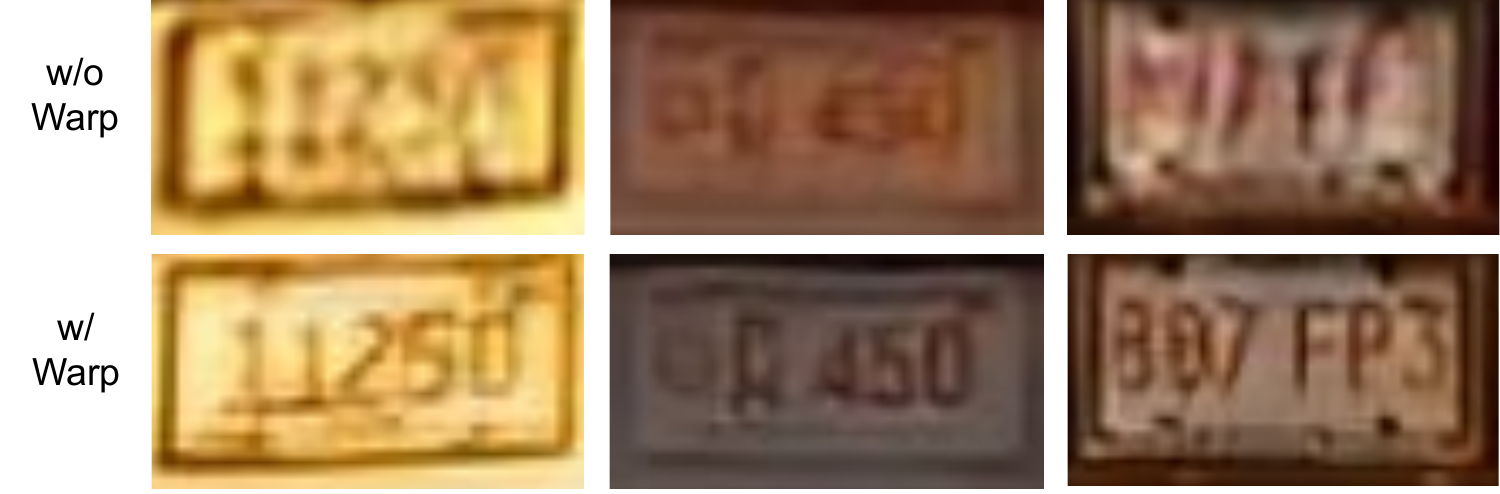}
     \caption{\textbf{Warping preserves fine details better than no-warp img2img-turbo~\cite{parmar2024one}.} License plate crops show that w/ Warp retains legible characters, while w/o Warp produces blurry, unreadable text.}
\end{figure}



\clearpage
\section{Ours vs. ChatGPT img2img}
\label{supsec:ours_chatgpt}

\begin{figure}[ht]
\centering
\includegraphics[width=0.77\linewidth]{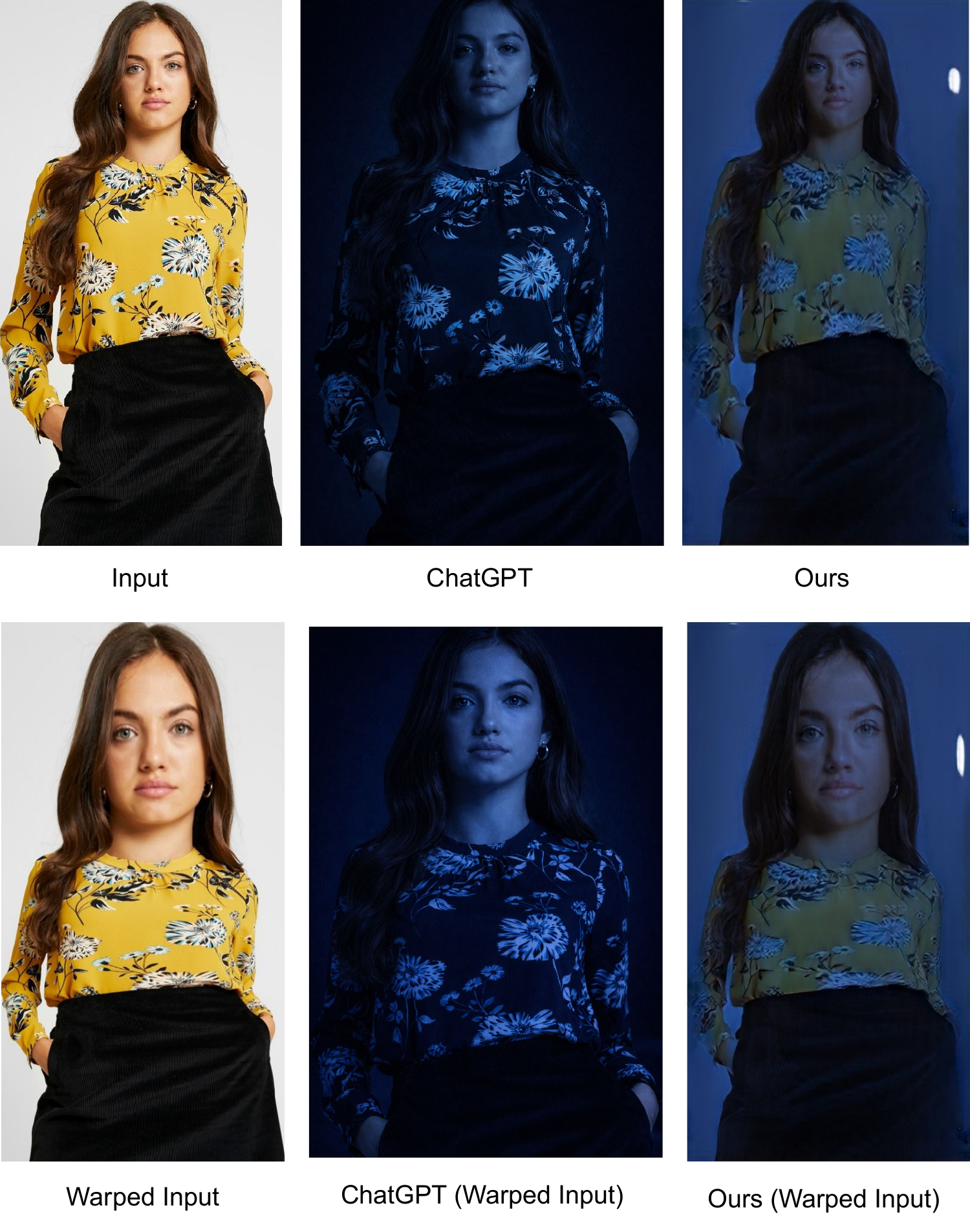}
\caption{\textbf{Our method vs. ChatGPT img2img for human relighting (moonlight).} \textit{Top}: ChatGPT changes face identity, as shown by different facial features after relighting, and also changes clothing identity, turning the golden-yellow cloth into a dark navy blue color. \textit{Bottom}: ChatGPT fails to handle warped inputs. While our method operates on warped images while preserving their scale, ChatGPT alters the spatial scale and structure.}
\label{fig:sup_chatgpt}
\end{figure}

\clearpage
\section{ChatGPT Annotation on Images}
\label{supsec:chatgpt_anno}

We use ChatGPT to generate high-quality image annotations for the outpainting and text-to-image (T2I) steps in our synthetic data generation pipeline. The  prompt template used for annotation is shown below.

\smallskip
\smallskip
\noindent ``Describe the person in the image using this exact format: Woman/Man, <pose>, wearing <top description>, paired with <bottom description if any>, <accessories if any>, <hair>, <expression if visible>, background of <scene and lighting>.''

\begin{figure*}[ht]
\centering

\begin{subfigure}[t]{0.48\linewidth}
    \centering
    \includegraphics[width=\linewidth]{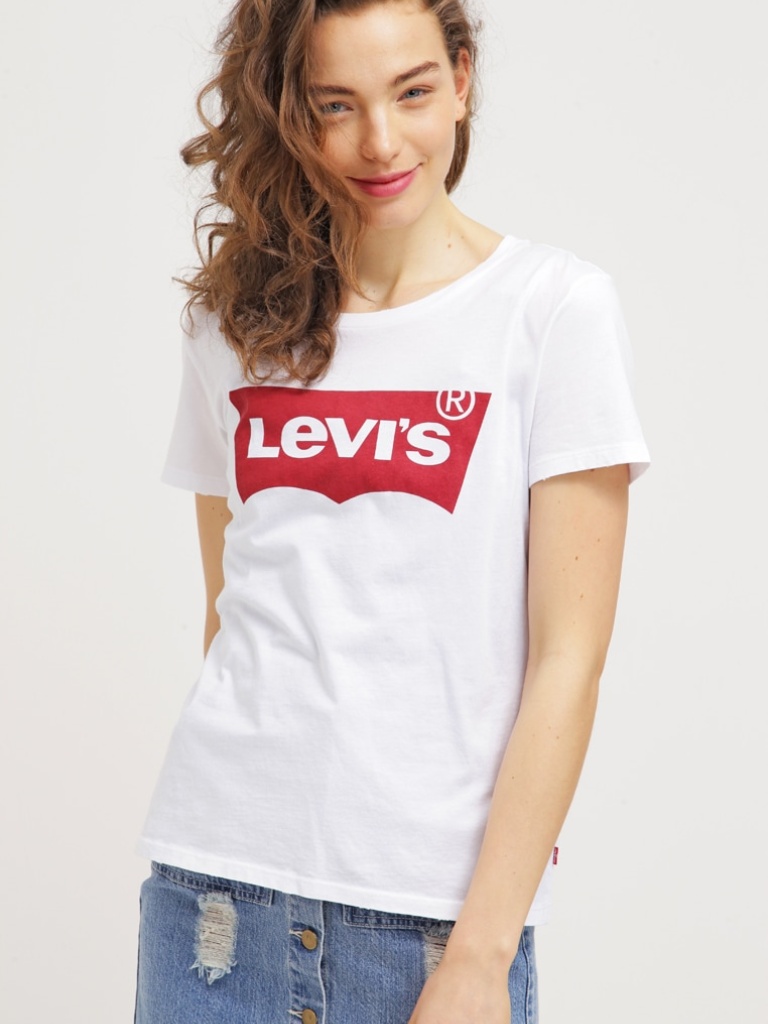}
    \caption{Woman, standing with body slightly turned and right arm behind the back, wearing a white short-sleeve t-shirt with a bold red Levi's® logo on the chest, paired with a high-waisted blue denim skirt featuring distressed patches and front buttons, no visible accessories, long wavy brown hair styled to one side, subtle smile with a relaxed expression, background of a plain light grey studio with soft even lighting.}
\end{subfigure}
\hfill
\begin{subfigure}[t]{0.48\linewidth}
    \centering
    \includegraphics[width=\linewidth]{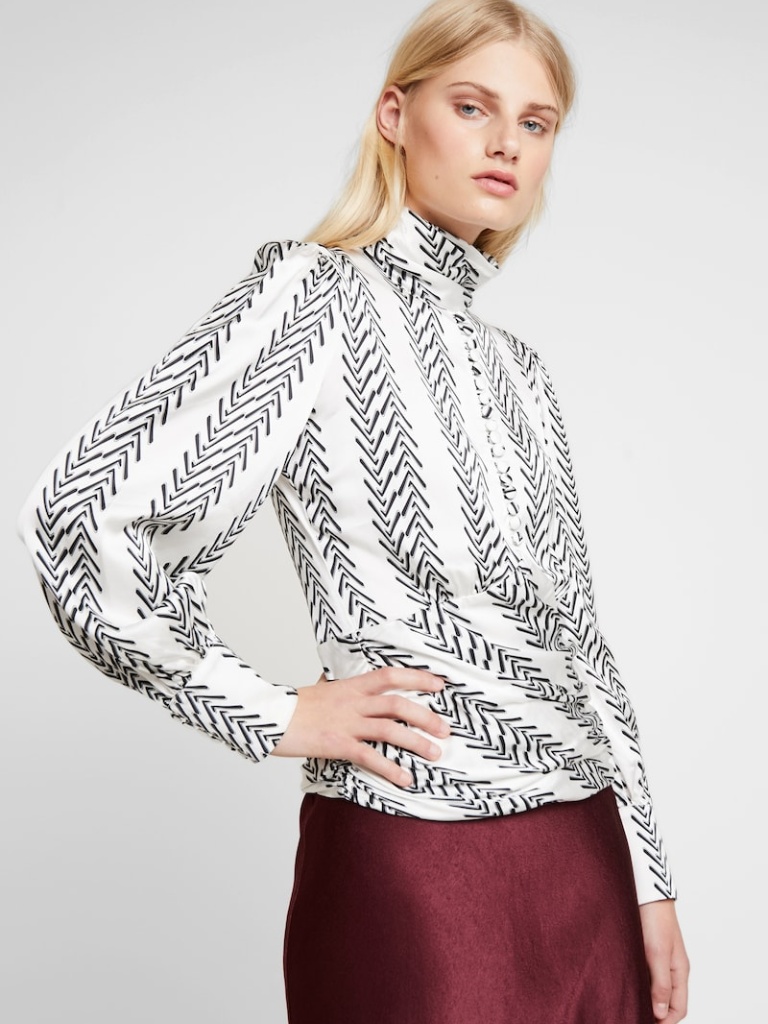}
    \caption{Woman, standing in a slight side pose with one hand on hip, wearing a high-neck long-sleeve blouse with black chevron print on white fabric and buttoned front, paired with a high-waisted burgundy satin skirt, no visible accessories, long straight blonde hair worn down, composed expression with subtle confidence, background of a plain light grey studio with soft lighting.}
\end{subfigure}

\caption{\textbf{Examples of ChatGPT-generated captions following our prompt template.}}
\label{fig:chatgpt_annotation_examples_1}
\end{figure*}

\begin{figure*}[t]
\centering

\begin{subfigure}[t]{0.48\linewidth}
    \centering
    \includegraphics[width=\linewidth]{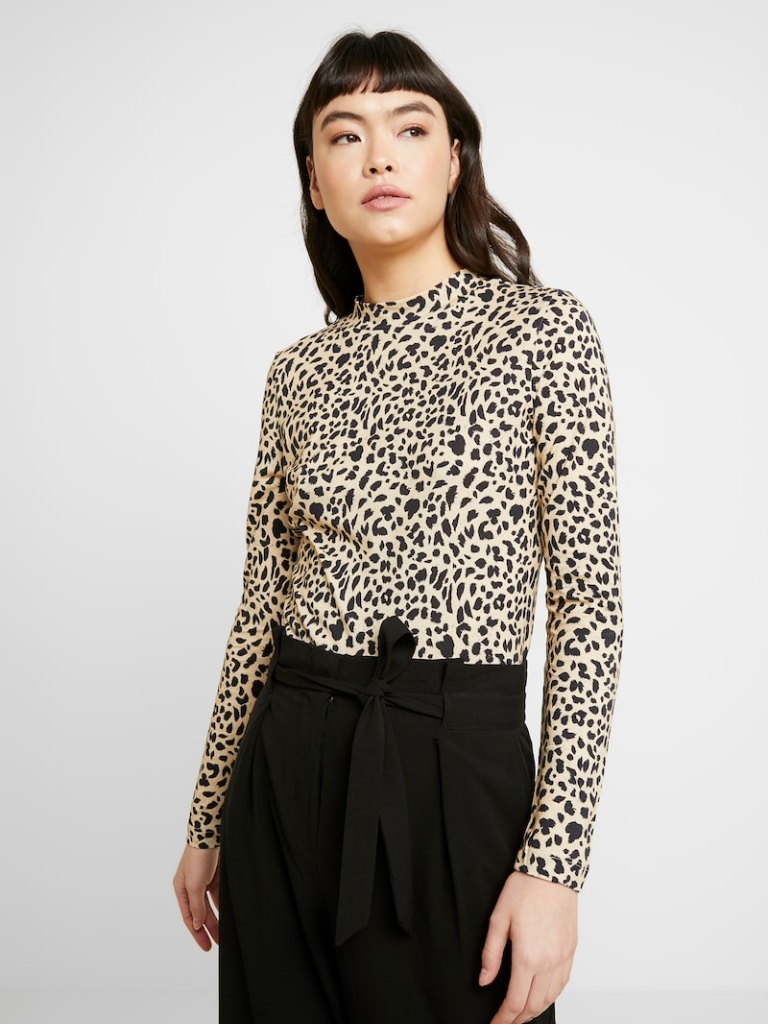}
    \caption{Woman, standing with right hand relaxed and head turned slightly to the side, wearing a beige long-sleeve top with black leopard print, paired with high-waisted black trousers featuring a tied waist belt, no visible accessories, long dark hair with bangs, calm and composed expression, background of a plain light grey studio with soft lighting.}
\end{subfigure}
\hfill
\begin{subfigure}[t]{0.48\linewidth}
    \centering
    \includegraphics[width=\linewidth]{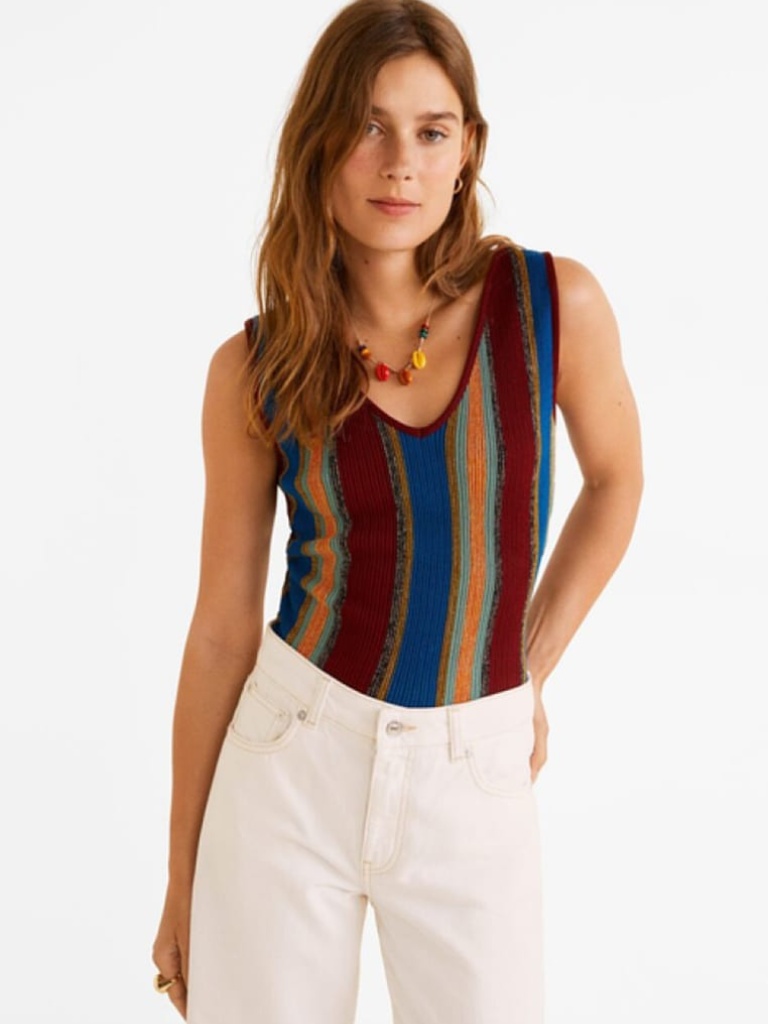}
    \caption{Woman, standing with left hand in pocket and body slightly angled, wearing a sleeveless multicolor ribbed knit top with vertical stripes in burgundy, blue, mustard, and teal, paired with high-waisted off-white jeans, accessorized with a colorful beaded necklace and small hoop earrings, long wavy brown hair worn loose, soft and confident expression, background of a plain white studio with soft lighting.}
\end{subfigure}

\caption{\textbf{Additional examples of ChatGPT-generated captions following our prompt template.}}
\label{fig:chatgpt_annotation_examples_2}
\end{figure*}

\smallskip
\smallskip
\noindent Examples of ChatGPT annotations are shown in Figure~\ref{fig:chatgpt_annotation_examples_1} and Figure~\ref{fig:chatgpt_annotation_examples_2}.

\clearpage

\section{ChatGPT Filtering Bad Generated Pairs}
\label{supsec:chatgpt_filtering}

We use ChatGPT to filter the bad generated pairs from our proposed Synthetic Paired Data Generation pipeline.

\subsection{ChatGPT Prompt Instructions}
\smallskip
\noindent ``This is a horizontal strip of 3 images:

\smallskip
\noindent - left: original image

\smallskip
\noindent - middle: generated base image

\smallskip
\noindent - right: generated relit image

\smallskip
\noindent Instructions:

\smallskip
\noindent 1. Does the \textbf{right} image match the following lighting description? \\
``<lighting description>'' \\
(Yes/No)

\smallskip
\noindent 2. Do both the \textbf{middle} and \textbf{right} images contain exactly one person? \\
(Yes/No)

\smallskip
\noindent 3. Same person? \\
(Yes/No)

\smallskip
\noindent 4. Same clothing? \\
(Yes/No)

\smallskip
\noindent 5. Same pose? \\
(Yes/No)

\smallskip
\noindent \textbf{FINAL ANSWER:} Yes (only if all answers are Yes). Otherwise, write No.''

\subsection{Evaluating Synthetic (T2I) Pairs} 

\begin{figure}[htbp]
    \centering
    \includegraphics[width=\linewidth]{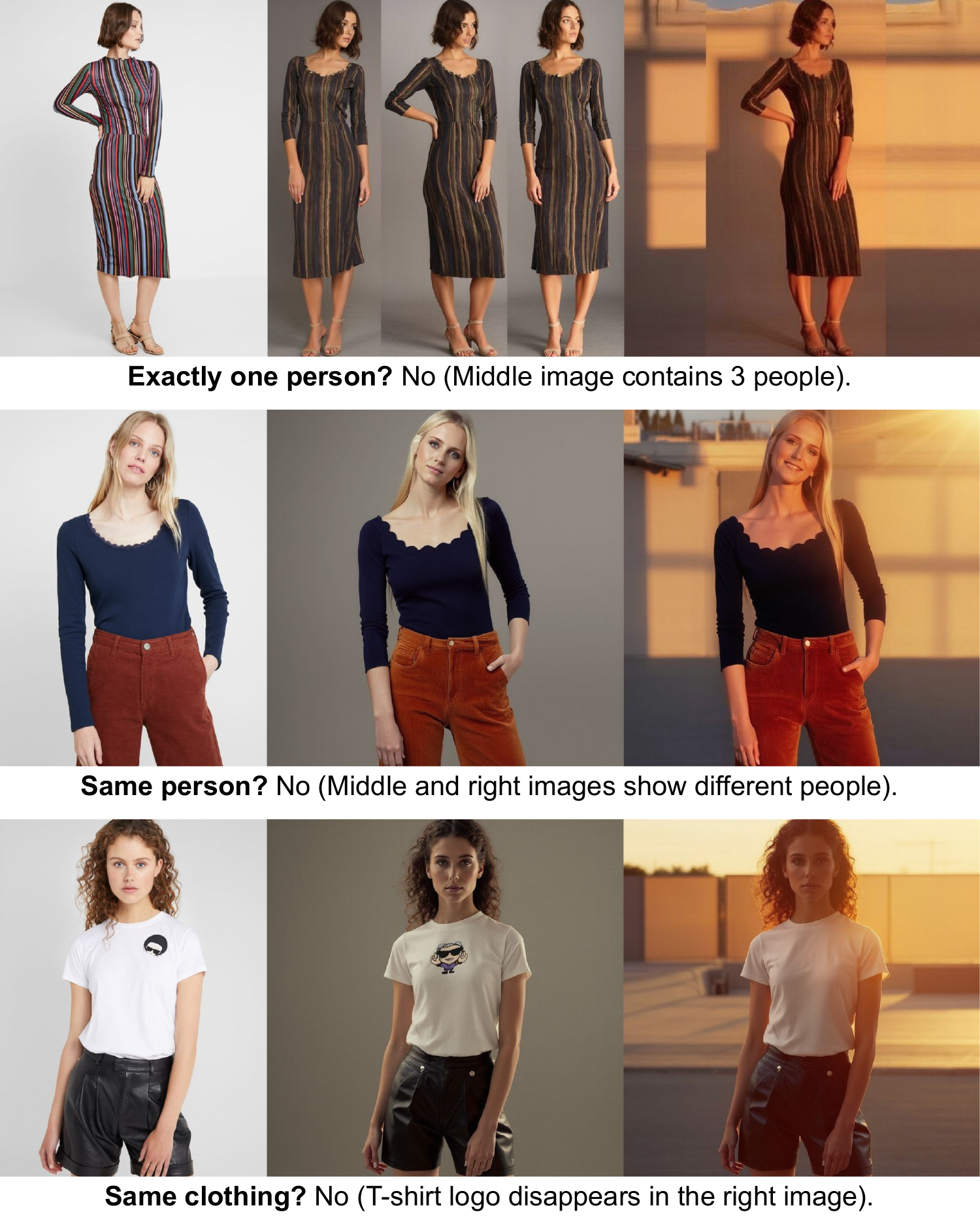}
    \caption{\textbf{Bad generated pairs filtered by ChatGPT.} \textit{Left}: original image. \textit{Middle}: generated base image. \textit{Right}: generated relit image. The bold questions below each panel evaluate the middle and right images.}
    \label{fig:sup_chatgpt_filter}
\end{figure}

Almost no images fail the \textit{lighting description} test. About 1\% fail the \textit{exactly one person} check, 1\% fail the \textit{same person} check, and 2\% fail the \textit{same clothing} check. Overall, 96\% of the generated pairs can be directly used for training, indicating that our pipeline is highly efficient. Examples of bad generated pairs filtered by ChatGPT are shown in Figure~\ref{fig:sup_chatgpt_filter}.

\subsection{Evaluating Human Relighting (I2I) Results}
We adapt the ChatGPT prompt instruction to evaluate human relighting (I2I) results by removing the leftmost image in the panel. The input image is placed in the middle position and the relit output image is placed on the right. We then use the ChatGPT API to evaluate whether the generated results satisfy several criteria (person identity, clothing identity, pose identity, and lighting faithfulness). The percentage of images passing each test is reported in Table~\ref{tab:chatgpt_api_relight}. Note that we omit the exactly one person check in this evaluation because all tested relighting results satisfy this condition.

\begin{table}[htbp]
\setlength{\tabcolsep}{6pt}
\centering
\caption{\textbf{ChatGPT API evaluation of relighting results.} We use the same filtering instruction to measure the percentage of generated images that pass each test (person identity, clothing identity, pose identity, and lighting faithfulness). Pass rates are averaged over four lighting prompts: golden sunlight, noon sunlight, moonlight, and foggy conditions. Our method outperforms baseline methods IC-Light~\cite{zhang2025scaling} and DreamLight~\cite{liu2026dreamlight}, and both BPS and warping further improve performance.}
\resizebox{\linewidth}{!}{
\begin{tabular}{l c c c c} 
\toprule
\textbf{Methods} & \begin{tabular}{c}\textbf{Person}\\\textbf{Identity}\end{tabular} & \begin{tabular}{c}\textbf{Cloth}\\\textbf{Identity}\end{tabular} & \begin{tabular}{c}\textbf{Pose}\\\textbf{Identity}\end{tabular} & \begin{tabular}{c}\textbf{Lighting}\\\textbf{Faithfulness}\end{tabular} \\
\midrule
IC-Light & 41.3 & 50.8 & 85.0 & 63.0 \\  
DreamLight & 53.8 & 60.3 & 87.5 & 63.8 \\  
\midrule
Ours (no BPS) & 73.0 & 88.3 & 92.0 & 83.5 \\  
Ours (no warp) & 76.5 & 90.5 & 91.5 & 86.5 \\  
\textbf{Ours} & \textbf{90.5} & \textbf{95.5} & \textbf{95.0} & \textbf{93.3} \\ 
\bottomrule
\end{tabular}
}
\label{tab:chatgpt_api_relight}
\end{table}

\section{Relighting Text Prompts}
\label{supsec:text_prompts}

Our detailed relight prompts as shown below. 

\noindent \textbf{noon sunlight}: `Relit with bright noon sunlight in a clear outdoor setting, casting soft natural shadows and illuminating the scene with crisp white daylight to create a clean, vibrant daytime atmosphere.'

\smallskip
\noindent \textbf{golden sunlight}: `Relit with warm golden sunlight during late afternoon, casting gentle directional shadows and bathing the scene in soft amber tones to create a calm, radiant atmosphere.'

\smallskip
\noindent \textbf{foggy}: `Relit in a dense foggy outdoor setting, producing soft diffused shadows and pale gray illumination that creates a quiet, atmospheric mood.'

\smallskip
\noindent \textbf{moonlight}: `Relit with cold moonlight in a minimalist nighttime scene, casting soft crisp shadows and bathing the scene in icy blue highlights to create a tranquil, distant atmosphere.'

\clearpage

\section{Additional Details: User Study}
\label{supsec:user_study_details}

As shown in the user study details in Table~\ref{tab:human_relighting_ci} and Table~\ref{tab:streettryon_relighting_ci}, our method achieves higher mean ratings with lower std and narrower CI bounds, indicating more consistent performance.

\begin{table}[ht]
\setlength{\tabcolsep}{4pt}
\caption{\textbf{User study on relighting (VITON-HD) with mean $\pm$ std [95\% CI].}}
\resizebox{\linewidth}{!}{
\begin{tabular}{lcccc}
\toprule
\textbf{Methods}
& \textbf{Person Identity}$\uparrow$
& \textbf{Cloth Identity}$\uparrow$
& \textbf{Image Quality}$\uparrow$
& \textbf{Lighting Faithfulness}$\uparrow$ \\
\midrule
IC-Light~\cite{zhang2025scaling}
& 3.4 $\pm$ 1.6 [3.2, 3.6] & 3.5 $\pm$ 1.6 [3.3, 3.7] & 3.5 $\pm$ 1.6 [3.3, 3.7] & 3.5 $\pm$ 1.6 [3.3, 3.7] \\
DreamLight~\cite{liu2026dreamlight}
& 3.5 $\pm$ 1.5 [3.3, 3.7] & 3.5 $\pm$ 1.6 [3.3, 3.7] & 3.5 $\pm$ 1.5 [3.3, 3.7] & 3.4 $\pm$ 1.6 [3.2, 3.6] \\
\midrule
Ours (no BPS)
& 3.7 $\pm$ 1.4 [3.5, 3.9] & 3.7 $\pm$ 1.4 [3.5, 3.9] & 3.6 $\pm$ 1.4 [3.4, 3.8] & 3.6 $\pm$ 1.4 [3.5, 3.8] \\
Ours (no Warp)
& 3.6 $\pm$ 1.5 [3.4, 3.8] & 3.6 $\pm$ 1.4 [3.4, 3.8] & 3.5 $\pm$ 1.5 [3.3, 3.7] & 3.5 $\pm$ 1.5 [3.4, 3.7] \\
\textbf{Ours}
& \textbf{4.4 $\pm$ 1.1 [4.2, 4.5]} & \textbf{4.4 $\pm$ 1.0 [4.3, 4.6]} & \textbf{4.3 $\pm$ 1.1 [4.2, 4.5]} & \textbf{4.2 $\pm$ 1.2 [4.1, 4.4]} \\
\bottomrule
\end{tabular}
}
\label{tab:human_relighting_ci}
\end{table}

\begin{table}[ht]
\setlength{\tabcolsep}{4pt}
\caption{\textbf{User study on relighting  (StreetTryOn) with mean $\pm$ std [95\% CI]. }}
\resizebox{\linewidth}{!}{
\begin{tabular}{lcccc}
\toprule
\textbf{Methods}
& \textbf{Person Identity}$\uparrow$
& \textbf{Cloth Identity}$\uparrow$
& \textbf{Image Quality}$\uparrow$
& \textbf{Lighting Faithfulness}$\uparrow$ \\
\midrule
IC-Light~\cite{zhang2025scaling}
& 3.3 $\pm$ 1.7 [3.1, 3.5] & 3.2 $\pm$ 1.6 [3.0, 3.4] & 3.3 $\pm$ 1.6 [3.0, 3.5] & 3.3 $\pm$ 1.6 [3.1, 3.5] \\
DreamLight~\cite{liu2026dreamlight}
& 3.2 $\pm$ 1.7 [2.9, 3.4] & 3.1 $\pm$ 1.7 [2.9, 3.3] & 3.1 $\pm$ 1.7 [2.9, 3.3] & 3.1 $\pm$ 1.7 [2.9, 3.3] \\
\midrule
Ours (no BPS)
& 3.6 $\pm$ 1.5 [3.4, 3.8] & 3.6 $\pm$ 1.4 [3.4, 3.8] & 3.5 $\pm$ 1.5 [3.3, 3.7] & 3.5 $\pm$ 1.5 [3.3, 3.7] \\
Ours (no Warp)
& 3.6 $\pm$ 1.5 [3.4, 3.8] & 3.6 $\pm$ 1.4 [3.4, 3.8] & 3.5 $\pm$ 1.5 [3.3, 3.7] & 3.6 $\pm$ 1.5 [3.4, 3.7] \\
\textbf{Ours}
& \textbf{4.3 $\pm$ 1.1 [4.2, 4.5]} & \textbf{4.3 $\pm$ 1.1 [4.1, 4.4]} & \textbf{4.2 $\pm$ 1.2 [4.0, 4.3]} & \textbf{4.1 $\pm$ 1.3 [4.0, 4.3]} \\
\bottomrule
\end{tabular}
}
\label{tab:streettryon_relighting_ci}
\end{table}

\clearpage
\section{Additional Results: Human Relighting}
\label{supsec:add_res_human}

\begin{figure}[ht]
\centering
\includegraphics[width=0.93\linewidth]{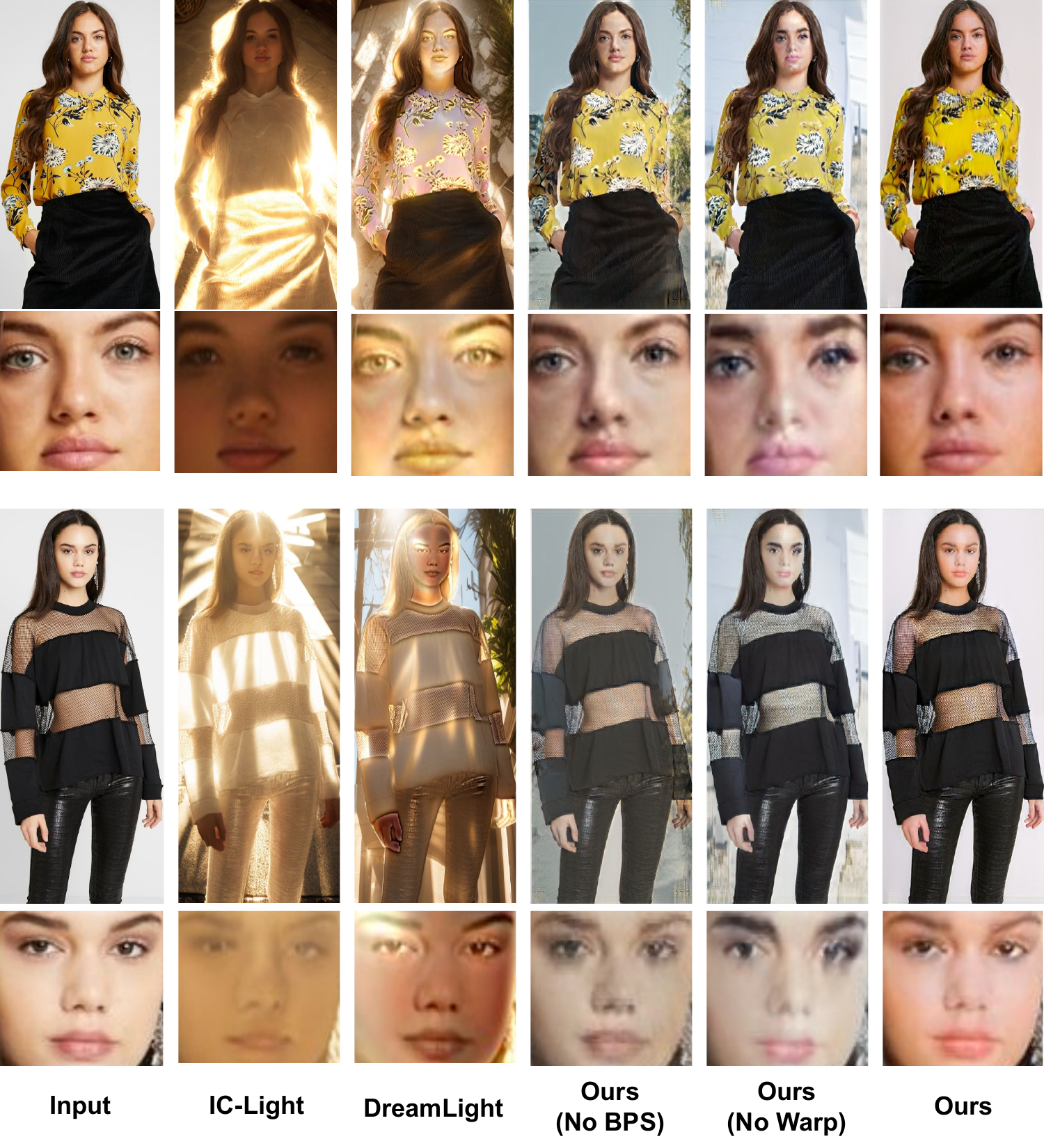}
\caption{\textbf{Additional qualitative comparison on human relighting (noon sunlight)}. }
\label{fig:vis_viton_supp_noon}
\end{figure}

\begin{figure}[t]
\centering
\includegraphics[width=0.93\linewidth]{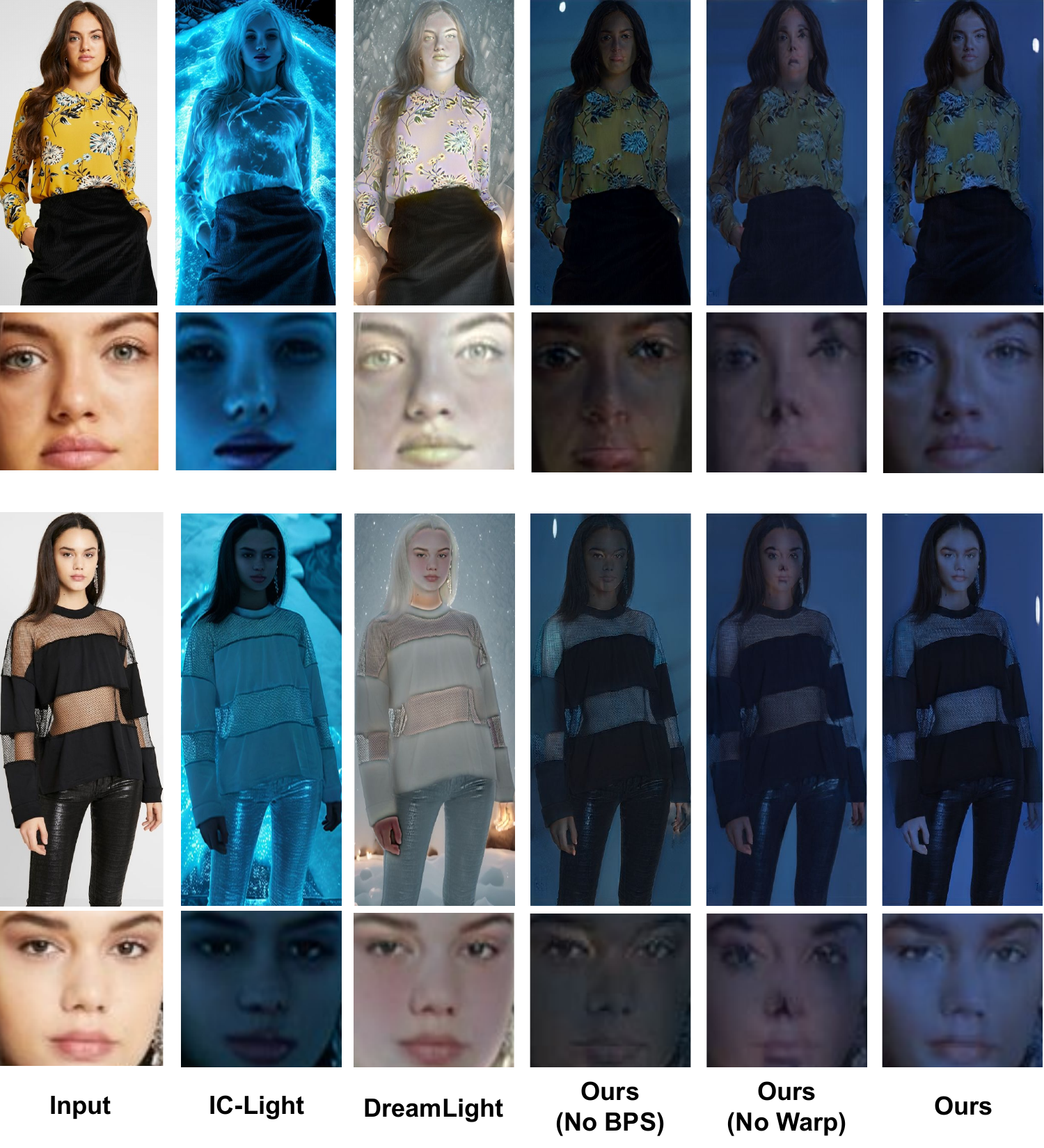}
\caption{\textbf{Additional qualitative comparison on human relighting (moonlight)}.}
\label{fig:vis_viton_supp_moon}
\end{figure}

\begin{figure}[t]
\centering
\includegraphics[width=0.93\linewidth]{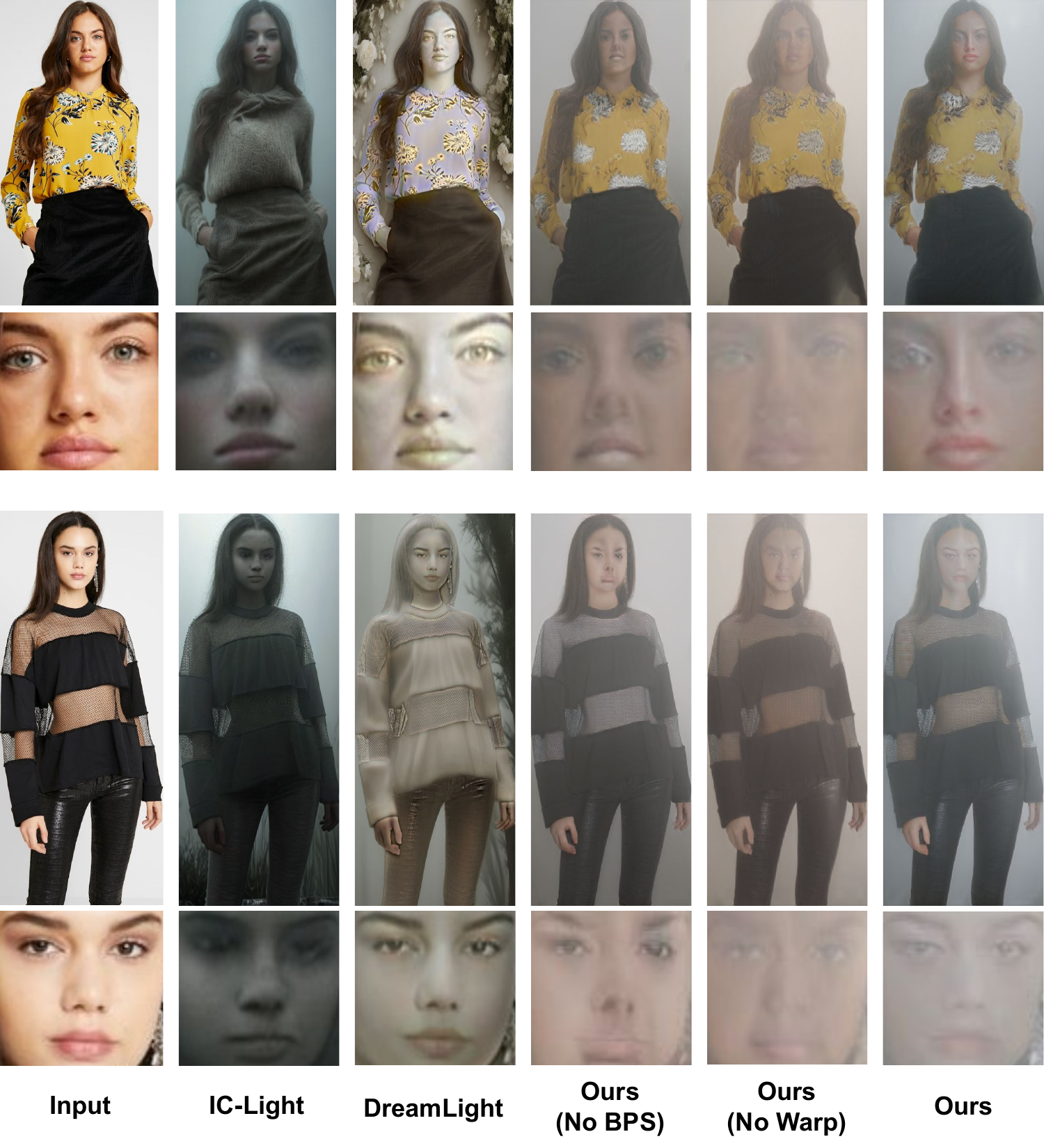}
\caption{\textbf{Additional qualitative comparison on human relighting (foggy)}.}
\label{fig:vis_viton_supp_foggy}
\end{figure}

\begin{figure}[t]
\centering
\includegraphics[width=0.93\linewidth]{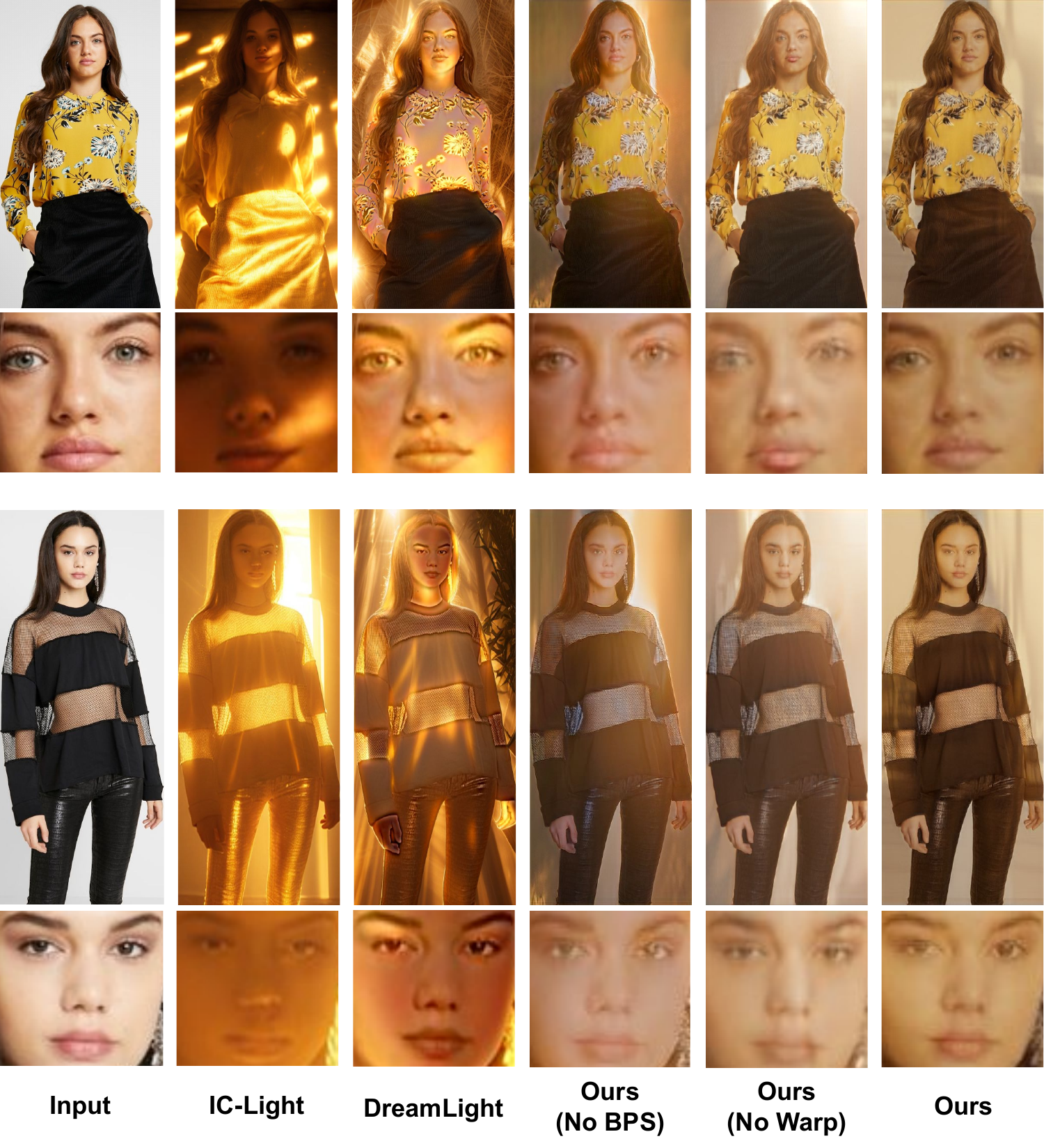}
\caption{\textbf{Additional qualitative comparison on human relighting (golden sunlight)}.}
\label{fig:vis_viton_supp_golden}
\end{figure}

We provide additional qualitative comparisons for human relighting in Figures~\ref{fig:vis_viton_supp_noon} (Noon Sunlight), \ref{fig:vis_viton_supp_moon} (Moonlight), \ref{fig:vis_viton_supp_foggy} (Foggy), and \ref{fig:vis_viton_supp_golden} (Golden Sunlight) on VITON-HD.


We notice that IC-Light~\cite{zhang2025scaling} and DreamLight~\cite{liu2026dreamlight} often struggle to preserve clothing identity and produce lighting effects that are less faithful to the relighting prompts. IC-Light also alter person identity due to distorted facial features. In contrast, our method, with the proposed BPS and warping, better preserves identity and clothing details while producing lighting effects that more faithfully match the prompts, leading to improved image quality.


\clearpage
\section{Additional Results: Driving Scene Relighting}
\label{supsec:add_res_driving}

\begin{figure}
\centering
\includegraphics[width=0.86\linewidth]{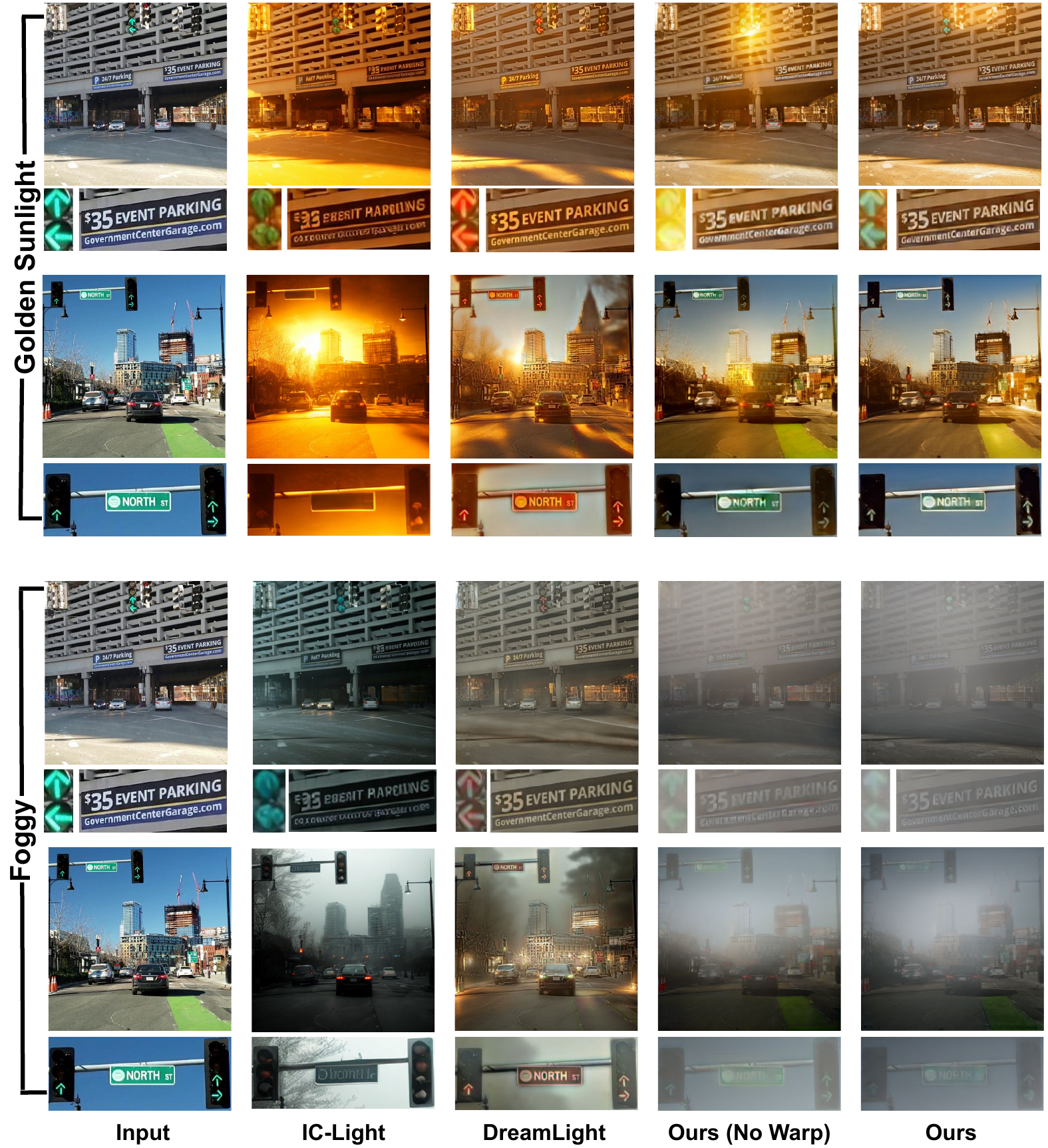}
\caption{\textbf{Additional qualitative comparison on driving scene relighting.} IC-Light~\cite{zhang2025scaling} and DreamLight~\cite{liu2026dreamlight} often produce lower image quality, less faithful lighting, and inconsistent semantics. For example, IC-Light turns green traffic arrows into circular lights or removes them, while DreamLight changes green arrows into red or yellow ones. In contrast, our method preserves semantic consistency and produces more realistic illumination. With our warping, small details such as traffic arrows and text on signs (e.g., `NORTH ST' and `GovernmentCenterGarage.com') become more legible.}
\label{fig:vis_roadwork_merged_supp}
\end{figure}

\clearpage
\section{Additional Results: Weather and Time-of-Day Driving Scene Translation}
\label{supsec:add_res_weather}

\begin{table}[ht]
\setlength{\tabcolsep}{4pt}
\caption{\textbf{Additional quantitative results on weather and time-of-day driving scene
translation.} KID is reported $\times$1000 and DINO-Struct $\times$100. }
\resizebox{\linewidth}{!}{
\begin{tabular}{lcccccccc}
\toprule
 & \multicolumn{4}{c}{\textbf{Cityscapes $\rightarrow$ Dense Foggy}} 
 & \multicolumn{4}{c}{\textbf{BDD100K (Rainy $\rightarrow$ Clear)}} \\
\cmidrule(lr){2-5}\cmidrule(lr){6-9}
Methods 
& FID$\downarrow$ & KID$\downarrow$ & \begin{tabular}[c]{@{}c@{}}Clean-\\FID$\downarrow$\end{tabular} & \begin{tabular}[c]{@{}c@{}}DINO\\Struct.$\downarrow$\end{tabular} 
& FID$\downarrow$ & KID$\downarrow$ & \begin{tabular}[c]{@{}c@{}}Clean-\\FID$\downarrow$\end{tabular} & \begin{tabular}[c]{@{}c@{}}DINO\\Struct.$\downarrow$\end{tabular} \\
\midrule
\begin{tabular}[c]{@{}l@{}}CycleGAN-Turbo\\(No Warp)\end{tabular} 
& 280.1 & 293.9 & 298.6 & \textbf{1.70} 
& 66.5  & 20.6  & 65.0  & 0.88 \\
\hdashline
\begin{tabular}[c]{@{}l@{}}CycleGAN-Turbo\\(With Warp)\end{tabular} 
& \textbf{238.7} & \textbf{234.5} & \textbf{263.1} & 2.38 
& \textbf{54.7} & \textbf{9.7} & \textbf{52.7} & \textbf{0.77} \\
\bottomrule
\end{tabular}
}
\label{tab:warp_ablation_results}
\end{table}

Additional quantitative results for weather and time-of-day driving scene translation, which we did not have space to include in the main paper, are shown in Table~\ref{tab:warp_ablation_results}. Adding our warping consistently improves FID~\cite{heusel2017gans}, KID~\cite{binkowski2018demystifying}, and Clean-FID~\cite{parmar2022aliased}, indicating better alignment with real images in the target domain.

\begin{figure}[t]
\centering
\includegraphics[width=\linewidth]{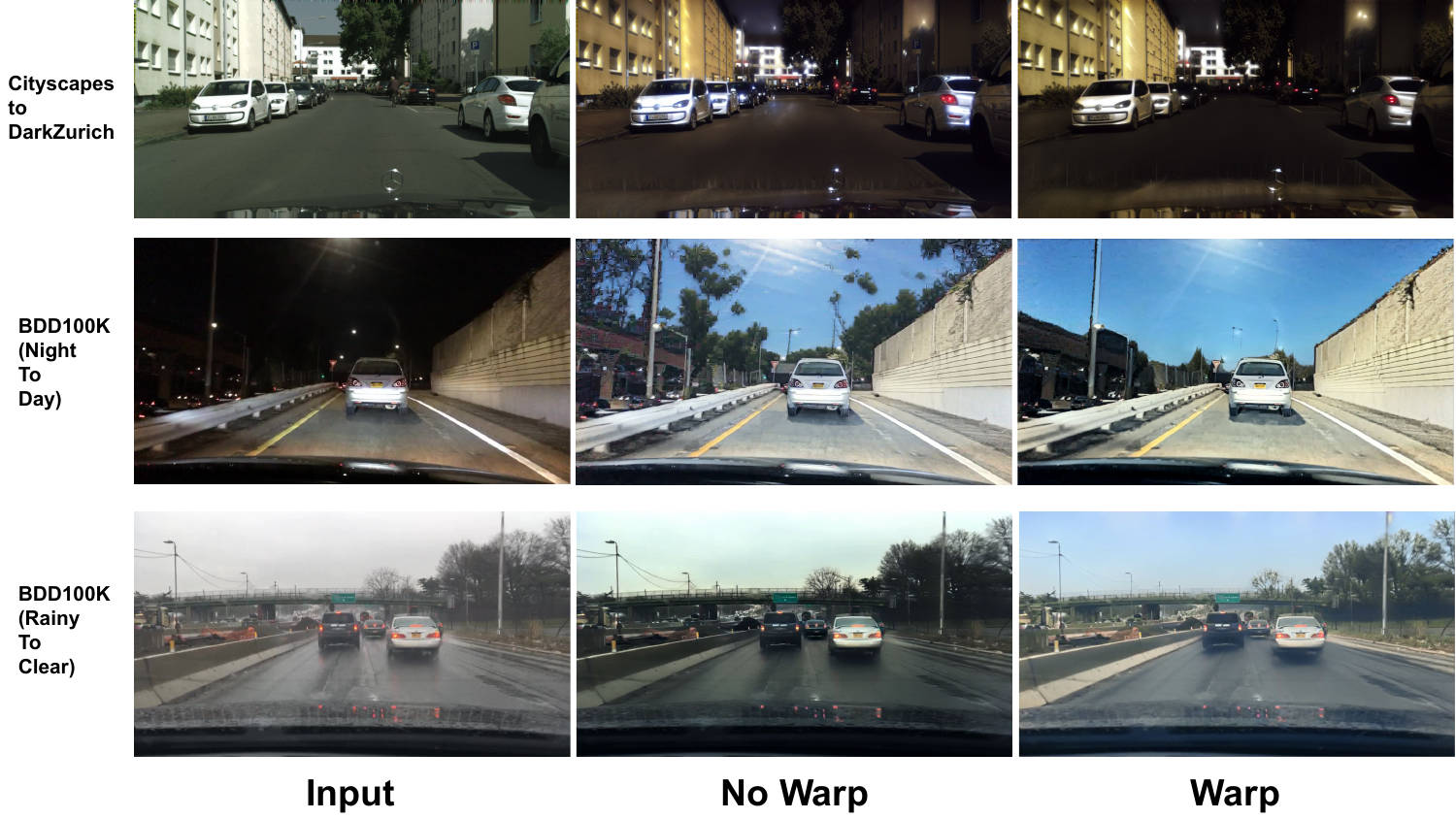}
\vspace{-0.5cm}
\caption{\textbf{Qualitative background improvement in weather and time-of-day driving scene translation} on multiple driving datasets, including Cityscapes~\cite{cordts2016cityscapes}, DarkZurich~\cite{sakaridis2019guided}, and BDD100K~\cite{yu2020bdd100k}. Compared with the baseline CycleGAN-Turbo~\cite{parmar2024one} without warping, adding our warping leads to more realistic illumination and contrast with fewer artifacts and distortions. While we have focused on warping and improving foreground objects, the background is also improved due to less area for potential hallucinations. Best viewed when zoomed in. }
\label{fig:vis_driving}
\end{figure}

\begin{figure}[t]
\centering
\includegraphics[width=\linewidth]{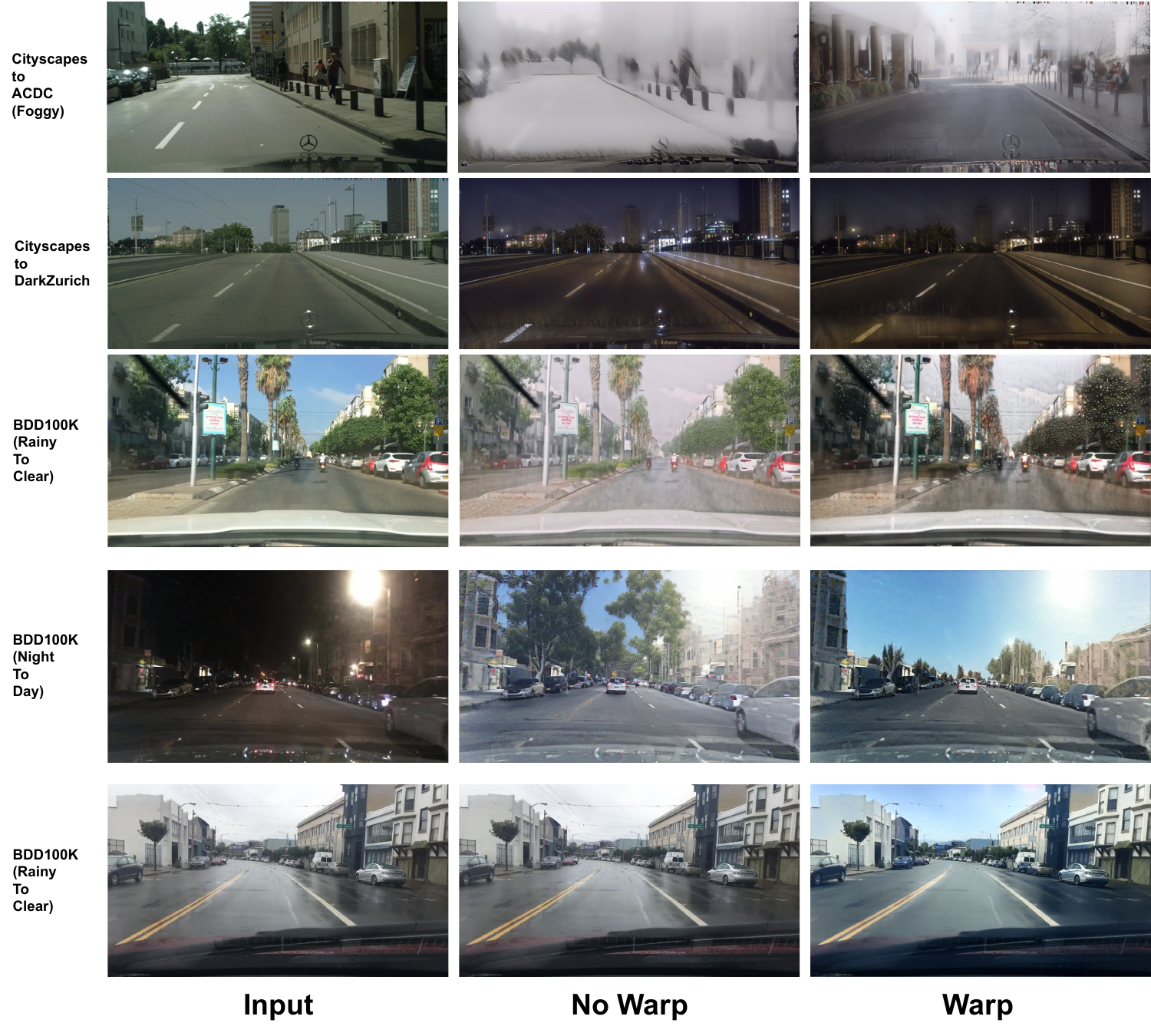}
\caption{\textbf{Additional qualitative background improvement in weather and time-of-day driving scene translation} on multiple driving datasets, including Cityscapes~\cite{cordts2016cityscapes}, ACDC~\cite{sakaridis2021acdc}, DarkZurich~\cite{sakaridis2019guided}, and BDD100K~\cite{yu2020bdd100k}. Across multiple weather and time-of-day translation tasks, adding our warping produces more realistic illumination and contrast with fewer artifacts and distortions compared with the baseline without warping. Although our warping is primarily designed to improve salient foreground regions, it also improves background appearance by reducing hallucinated content and producing more coherent global lighting and colors. Best viewed when zoomed in.}
\label{fig:vis_drive_i2i_supp}
\end{figure}

Additional qualitative comparisons are shown in Figures~\ref{fig:vis_driving} and ~\ref{fig:vis_drive_i2i_supp}. Our warping improves illumination consistency and reduces artifacts. While the warping is primarily designed to preserve foreground structures, it also improves the background by limiting the spatial regions where hallucinations may occur.

For the quantitative comparisons, we follow img2img-turbo~\cite{parmar2024one} and report DINO-Struct~\cite{tumanyan2022splicing} for completeness. Our warping improves DINO-Struct for BDD100K (Rainy $\rightarrow$ Clear). For Cityscapes $\rightarrow$ Dense Foggy, warping slightly degrades DINO-Struct, where the metric may be less reliable due to the reduced robustness of DINO features under severe weather conditions~\cite{zhou2022understanding,zhang2023tale}.

We notice that the training of CycleGAN-Turbo, based on the CycleGAN framework, is highly unstable~\cite{yuan2018unsupervised,zheng2023tpsence,wang2024cyclegan,parmar2024one}. For extreme weather-to-clear translation tasks, including Dense Fog $\rightarrow$ Cityscapes and ACDC Foggy $\rightarrow$ Cityscapes, which are ill-posed~\cite{chen2022learning,wang2025modem}, we noticed that the baseline model fails to converge with loss divergence, and warping alone cannot resolve this issue. Sophisticated models with greater capability are likely needed, which we leave for future work.

%
%

\end{document}